\documentclass[11pt]{article}

\usepackage[final]{acl}

\usepackage{times}
\usepackage{latexsym}

\usepackage[T1]{fontenc}

\usepackage[utf8]{inputenc}

\usepackage{microtype}

\usepackage{inconsolata}

\usepackage{graphicx}
\usepackage{booktabs}
\usepackage{lineno}
\usepackage{amssymb}
\usepackage{multirow}
\usepackage{makecell}
\usepackage{pifont}
\usepackage[svgnames]{xcolor}
\usepackage{enumitem}
\usepackage{xcolor}
\usepackage{colortbl}
\usepackage[most]{tcolorbox}
\usepackage{caption}
\usepackage{subcaption}
\usepackage{hyperref}
\usepackage{multirow}
\usepackage{listings}
\tcbuselibrary{breakable}

\newtcolorbox{domain_cases}[1][]{
    colback=teal!4,             
    colframe=teal!70!black,     
    boxrule=0pt,                
    leftrule=1.5pt,             
    rightrule=1pt,              
    sharp corners,              
    enhanced, 
    breakable, 
    break at=-0.5cm, 
    pad at break*=2mm, 
    toprule at break=0pt,       
    bottomrule at break=0pt, 
    topsep at break=0pt, 
    bottomsep at break=0pt, 
    #1 
}
\newtcolorbox{failure_cases}[2][]{
    colback=gray!5,             
    colframe=gray!60!black,     
    colbacktitle=gray!20,       
    title={#2}, 
    sharp corners, 
    boxrule=1pt, 
    titlerule=0pt, 
    fonttitle=\bfseries\color{black}, 
    enhanced, 
    breakable, 
    break at=-0.5cm, 
    pad at break*=2mm, 
    toprule at break=1pt, 
    bottomrule at break=1pt, 
    topsep at break=0pt, 
    bottomsep at break=0pt, 
    #1 
}

\newtcolorbox{llm_judge_search}[2][]{
    colback=blue!10, 
    colframe=blue!80!black, 
    colbacktitle=blue!50, 
    title={#2}, 
    sharp corners, 
    boxrule=1pt, 
    titlerule=0pt, 
    fonttitle=\bfseries\color{black}, 
    enhanced,                    
    breakable,                   
    break at=-0.5cm,              
    pad at break*=2mm,            
    toprule at break=1pt,         
    bottomrule at break=1pt,      
    topsep at break=0pt,          
    bottomsep at break=0pt,       
    #1 
}

\newtcolorbox{agent_judge_search}[2][]{
    colback=cyan!10, 
    colframe=cyan!80!black, 
    colbacktitle=cyan!50, 
    title={#2}, 
    sharp corners, 
    boxrule=1pt, 
    titlerule=0pt, 
    fonttitle=\bfseries\color{black}, 
    enhanced,                    
    breakable,                   
    break at=-0.5cm,              
    pad at break*=2mm,            
    toprule at break=1pt,         
    bottomrule at break=1pt,      
    topsep at break=0pt,          
    bottomsep at break=0pt,       
    #1 
}

\newtcolorbox{llm_judge_ds}[2][]{
    colback=orange!10, 
    colframe=orange!80!black, 
    colbacktitle=orange!50, 
    title={#2}, 
    sharp corners, 
    boxrule=1pt, 
    titlerule=0pt, 
    fonttitle=\bfseries\color{black}, 
    enhanced,                    
    breakable,                   
    break at=-0.5cm,              
    pad at break*=2mm,            
    toprule at break=1pt,         
    bottomrule at break=1pt,      
    topsep at break=0pt,          
    bottomsep at break=0pt,       
    #1 
}

\newtcolorbox{agent_judge_ds}[2][]{
    colback=red!10, 
    colframe=red!80!black, 
    colbacktitle=red!50, 
    title={#2}, 
    sharp corners, 
    boxrule=1pt, 
    titlerule=0pt, 
    fonttitle=\bfseries\color{black}, 
    enhanced,                    
    breakable,                   
    break at=-0.5cm,              
    pad at break*=2mm,            
    toprule at break=1pt,         
    bottomrule at break=1pt,      
    topsep at break=0pt,          
    bottomsep at break=0pt,       
    #1 
}

\newtcolorbox{llm_judge_gui}[2][]{
    colback=violet!10, 
    colframe=violet!80!black, 
    colbacktitle=violet!50, 
    title={#2}, 
    sharp corners, 
    boxrule=1pt, 
    titlerule=0pt, 
    fonttitle=\bfseries\color{black}, 
    enhanced,                    
    breakable,                   
    break at=-0.5cm,              
    pad at break*=2mm,            
    toprule at break=1pt,         
    bottomrule at break=1pt,      
    topsep at break=0pt,          
    bottomsep at break=0pt,       
    #1 
}

\newtcolorbox{agent_judge_gui}[2][]{
    colback=magenta!10, 
    colframe=magenta!80!black, 
    colbacktitle=magenta!50, 
    title={#2}, 
    sharp corners, 
    boxrule=1pt, 
    titlerule=0pt, 
    fonttitle=\bfseries\color{black}, 
    enhanced,                    
    breakable,                   
    break at=-0.5cm,              
    pad at break*=2mm,            
    toprule at break=1pt,         
    bottomrule at break=1pt,      
    topsep at break=0pt,          
    bottomsep at break=0pt,       
    #1 
}

\newcommand{\modelicon}[2]{%
  \raisebox{-0.25\height}{\includegraphics[height=1.15em]{figure/model_icon/#1}}~#2
}

\newcommand{\cmark}{\textcolor{green!60!black}{\ding{51}}} 
\newcommand{\xmark}{\textcolor{red!90!black}{\ding{55}}} 

\title{AJ-Bench: Benchmarking Agent-as-a-Judge for Environment-Aware Evaluation}

\author{
    \textbf{Wentao Shi$^{\clubsuit *}$, Yu Wang$^{\clubsuit *}$, Yuyang Zhao$^{\clubsuit *}$, Yuxin Chen$^{\diamondsuit}$, Fuli Feng$^{\clubsuit}$, Xueyuan Hao$^{\spadesuit}$,} \\
    \textbf{ Xi Su$^{\spadesuit}$, Qi Gu$^{\spadesuit \dagger}$, Hui Su$^{\spadesuit}$, Xunliang Cai$^{\spadesuit}$, Xiangnan He$^{\clubsuit \dagger}$} \\
    \vspace{0.2cm}
    $^{\clubsuit}$University of Science and Technology of China \ 
    $^{\diamondsuit}$National University of Singapore \ 
    $^{\spadesuit}$Meituan \\
    \texttt{\{shiwentao123, zhaoyuyang\}@mail.ustc.edu.cn, \{guqi03\}@meituan.com} \\
    \texttt{\{terencewang0809, fulifeng93\}@gmail.com}
}

\begin{document}
\maketitle

\begingroup
\renewcommand\thefootnote{}
\footnotetext{$^*$Equal contribution. $^\dagger$Corresponding authors.}
\addtocounter{footnote}{-1}
\endgroup

\begin{abstract}
As reinforcement learning continues to scale the training of large language model–based agents, reliably verifying agent behaviors in complex environments has become increasingly challenging. Existing approaches rely on rule-based verifiers or LLM-as-a-Judge models, which struggle to generalize beyond narrow domains. Agent-as-a-Judge addresses this limitation by actively interacting with environments and tools to acquire verifiable evidence, yet its capabilities remain underexplored.

We introduce a benchmark \textbf{AJ-Bench} to systematically evaluate Agent-as-a-Judge across three domains—search, data systems, and graphical user interfaces—comprising 155 tasks and 516 annotated trajectories. The benchmark comprehensively assesses judge agents’ abilities in information acquisition, state verification, and process verification. Experiments demonstrate consistent performance gains over LLM-as-a-Judge baselines, while also revealing substantial open challenges in agent-based verification. Our data and code are available at \href{https://aj-bench.github.io/}{https://aj-bench.github.io/}.
\end{abstract}

\section{Introduction}

\begin{quote}
\emph{"An intelligent system cannot be evaluated independently of the environment in which it operates."}
\end{quote}


%



The rapid progress of large language models (LLMs) has catalyzed the emergence of LLM-based agents that exhibit strong capabilities in long-horizon planning, multi-step reasoning, and tool use within complex environments~\citep{mohammadi2025evaluation, guo2024large, yao2022react}. To further advance agent performance across a diverse range of tasks, reinforcement learning (RL) plays a pivotal role by enabling agents to acquire more robust and transferable behaviors~\citep{chen2025reinforcement, cheng2025agent}. However, as RL computation continues to scale, a fundamental challenge emerges: how to verify agent behaviors in novel environments at scale.

\begin{figure}[t] 
    \centering
    \includegraphics[width=1.0\linewidth]{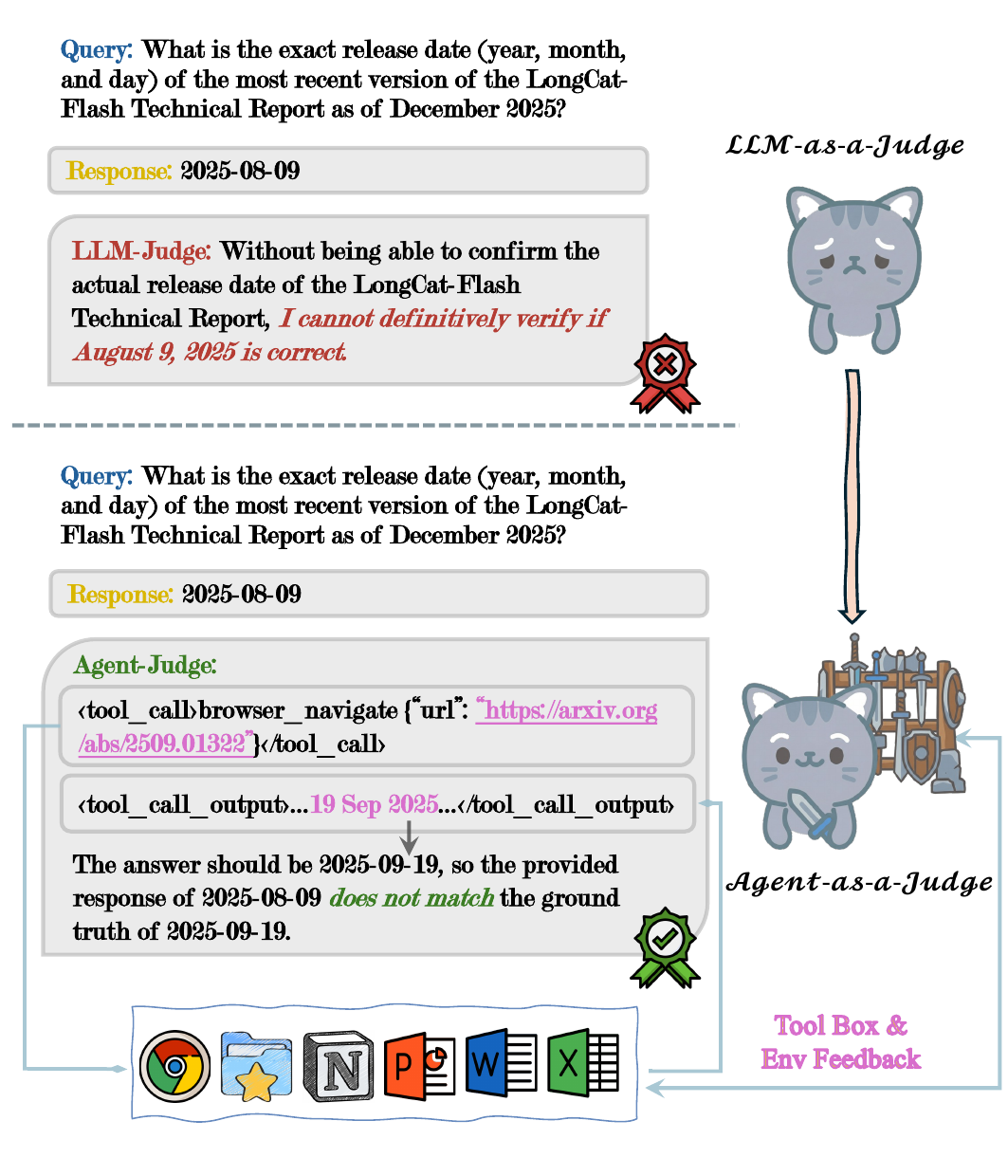} 
    \caption{Agent-as-a-Judge outperforms LLM-as-a-Judge by using tools and environment access to verify the correct release date.}
    \label{fig:intro_figure}
\end{figure}

\begin{table*}[t]
    \centering
    \begin{tabular}{lcccc}
        \toprule
        \textbf{Benchmark} 
        & \makecell{\textbf{Evaluation}\\\textbf{Target}} 
        & \makecell{\textbf{Multi}\\\textbf{Domain}}  
        & \textbf{Env-Aware} 
        & \makecell{\textbf{Agentic}\\\textbf{Interaction}} \\
        \midrule
        RewardBench~\cite{DBLP:conf/naacl/LambertPMMLCDKZCSH25} 
            & LLM-as-a-Judge & \cmark & \xmark & \xmark \\
        RM-Bench~\cite{DBLP:conf/iclr/Liu0M00L25}  
            & LLM-as-a-Judge & \cmark & \xmark & \xmark \\
        JudgeBench~\cite{DBLP:conf/iclr/TanZMTC0PS25}  
            & LLM-as-a-Judge & \cmark & \xmark & \xmark \\
        AgentRewardBench~\cite{DBLP:conf/acl/MenJC00025}
            & LLM-as-a-Judge & \cmark  &\xmark   &\xmark \\
        DevAI~\cite{DBLP:conf/icml/Zhuge0AWKXLCKTS25} 
            & Agent-as-a-Judge & \xmark & \cmark & \cmark \\
        \midrule
        AJ-Bench 
            & Agent-as-a-Judge & \cmark & \cmark & \cmark \\
        
        \bottomrule
    \end{tabular}
    \caption{
        Comparison of evaluation benchmarks for judges. \textbf{Multi-Domain} denotes coverage across multiple domains. \textbf{Env-Aware} indicates access to environment states. \textbf{Agentic Interaction} reflects whether active interaction (e.g., tool use) is allowed for judges.
    }
    \label{tab:rm_comparison}
\end{table*}

Recent studies primarily rely on rule-based verification~\citep{shao2024deepseekmath, mroueh2025reinforcement}, where agent trajectories are evaluated against predefined rules~\citep{wei2025gtr}. While effective for narrowly scoped tasks, such methods do not generalize to complex, realistic settings (e.g., scientific hypothesis verification or essay-level fact checking), where handcrafted rules are insufficient~\citep{huang2025automated}. In parallel, LLM-as-a-Judge approaches have been explored, but their judgements are ultimately grounded in surface-level textual signals~\citep{li2025generation, gu2025surveyllmasajudge}. To address these limitations, a natural progression is to endow the verifier with agency. By actively interacting with the environment, an agent-based judge can reproduce execution trajectories, verify intermediate states, and assess tool usage~\citep{DBLP:conf/icml/Zhuge0AWKXLCKTS25}.



Despite its conceptual appeal, the verification capability of Agent-as-a-Judge systems remains largely unexplored. While a small number of studies examine the consistency between Agent-as-a-Judge and human judgements on limited benchmarks~\citep{DBLP:conf/icml/Zhuge0AWKXLCKTS25}, these analyses are largely confined to small-scale datasets and narrow domains such as code verification, and therefore cannot offer a comprehensive assessment of Agent-as-a-Judge capability in open-ended settings. Meanwhile, these benchmarks fail to capture the more fundamental challenges faced by judge agents, including deciding when interaction is necessary, how to leverage tools effectively, and what constitutes sufficient and verifiable evidence for reliable judgement in open-ended environments.



In this work, we move toward a systematic evaluation of Agent-as-a-Judge as a distinct and general capability. We introduce a comprehensive benchmark \textbf{AJ-Bench} that explicitly requires judge agents to interact with environments and leverage external tools to obtain evidence beyond the given trajectories. The benchmark covers three domains—search, data system (DS), and graphical user interface (GUI)—and consists of 155 tasks spanning a wide range of complex agent behaviors. We further collect 516 trajectories annotated with binary (positive and negative) labels. Judge agents are evaluated using the F1 score between their predictions and the ground-truth annotations. The collected tasks and trajectories jointly assess key judging capabilities, including (i) information acquisition via external search, (ii) state verification through tool-assisted interaction with environments, and (iii) process verification by inspecting critical actions and execution steps. Our benchmark enables controlled comparisons between LLM-as-a-Judge and Agent-as-a-Judge, revealing clear qualitative differences in evaluation behavior. Agent-as-a-Judge consistently outperforms LLM-as-a-Judge, achieving an average improvement of 0.13 in F1. However, its absolute performance remains not yet saturated, with an average F1 score of 0.72, leaving substantial room for further improvement.


Our contributions can be summarized as follows:
\begin{itemize}[leftmargin=*]
    \item We introduce the first comprehensive benchmark AJ-Bench for evaluating Agent-as-a-Judge systems, enabling rich interaction with environments and systematic assessment of their judgement capabilities.
    \item We conduct systematic comparisons between Agent-as-a-Judge and LLM-as-a-Judge paradigms, demonstrating that equipping judging agents with tools and environment substantially improves judgement accuracy.
    \item Experiments across three representative domains indicate that, despite their advantages, judge agents still show imperfect performance in evaluating complex, multi-step behaviors, suggesting substantial headroom for future improvement.
\end{itemize}

\section{Related Work}
In this section, we briefly introduce benchmarks for evaluating LLM-based Judges (\S\ref{sec:relatedwork:benchmark llm judges}), the development of Agent-as-a-Judge (\S\ref{sec:relatedwork:agent-as-a-judge}), and benchmarks for evaluating Task-Solving Agents (\S\ref{sec:relatedwork:benchmark task-solving agent}).

\begin{figure*}[t] 
    \centering
    \includegraphics[width=1.0\linewidth]{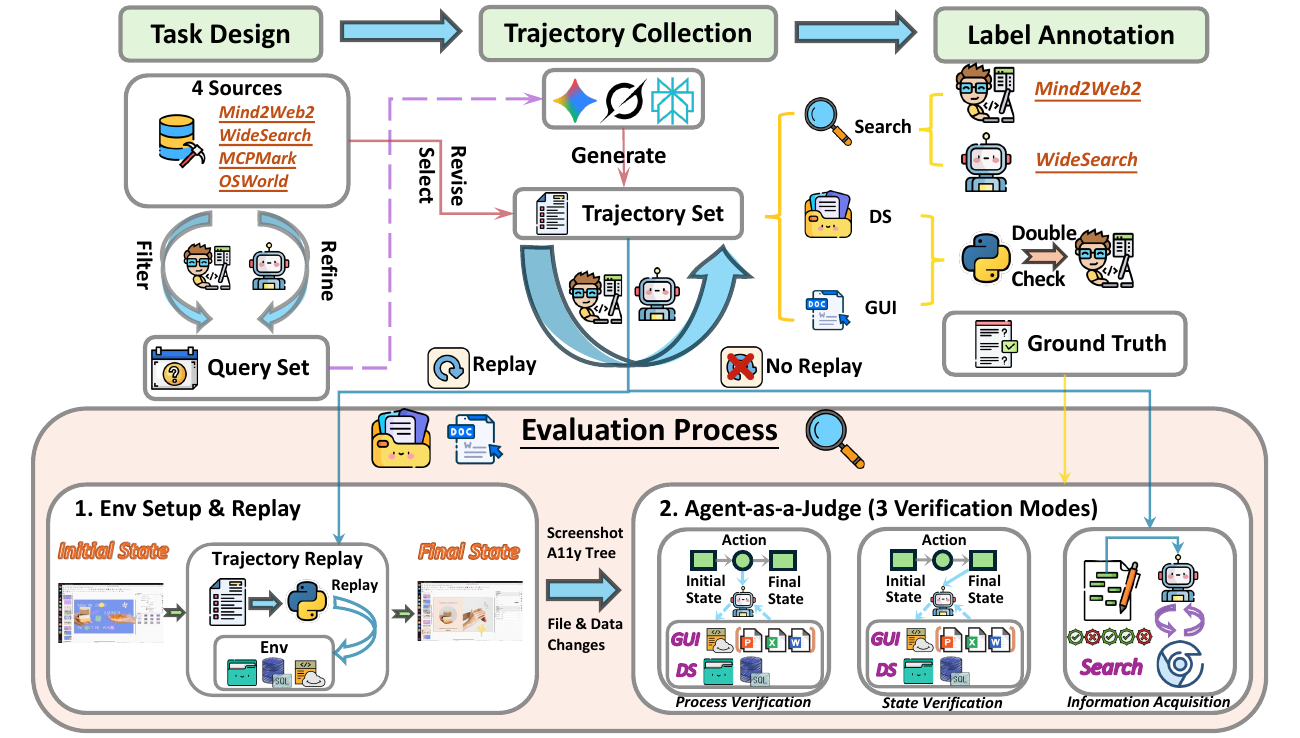} 
    \caption{\textbf{Overview of the benchmark and evaluation pipeline.} The upper part illustrates the benchmark construction process, including task design, trajectory collection, and label annotation. The lower part depicts the evaluation workflow of Agent-as-a-Judge, where the environment is initialized prior to evaluation.}
    \label{fig:method_figure}
\end{figure*}

\subsection{Benchmarks for LLM-Based Judges}\label{sec:relatedwork:benchmark llm judges}
To evaluate LLM-based judges, early benchmarks primarily measured the alignment between judges’ outputs and human judgements, emphasizing stylistic agreement over factual or logical correctness~\citep{DBLP:conf/nips/ZhengC00WZL0LXZ23, DBLP:conf/acl/WangLCCZLCKLLS24, DBLP:journals/corr/abs-2308-01862}. More recent work has shifted focus toward evaluating judges’ capacity to assess factual accuracy and reasoning, with datasets such as LLMBar~\citep{DBLP:conf/iclr/ZengYG0G024} targeting instruction-following tasks and JudgeBench~\citep{DBLP:conf/iclr/TanZMTC0PS25} focusing on reasoning. Comprehensive benchmarks like RewardBench~\citep{DBLP:conf/naacl/LambertPMMLCDKZCSH25} and RM-Bench~\citep{DBLP:conf/iclr/Liu0M00L25} further examine judges across diverse domains including safety, dialogue, and reasoning. Extending this line of research, AgentRewardBench~\citep{DBLP:conf/acl/MenJC00025} evaluate judges on agent trajectories, assessing their ability to judge task execution and planning. In contrast, our work investigates judges’ ability to evaluate agent behavior in interactive environments, requiring tool use and environment engagement, thereby introducing a more challenging setting.

\subsection{Agent-as-a-Judge}\label{sec:relatedwork:agent-as-a-judge}

Agent-as-a-Judge was initially introduced by~\citet{DBLP:conf/icml/Zhuge0AWKXLCKTS25} to incorporate agentic capabilities into verification. Recent work integrates external tool use (e.g., code execution, calculator) to enhance judges’ ability in reasoning tasks~\citep{han2025verifiagent, sung2025verilahumancenteredevaluationframework, sadhuka2025evaluatorreliableagentverifiers}, as demonstrated by Themis~\citep{DBLP:conf/iclr/0040CW0T0024}, TIR-Judge~\citep{DBLP:journals/corr/abs-2510-23038} and Agentic Reward Modeling~\citep{DBLP:conf/acl/0015QW000L25}. However, these efforts have largely remained confined to reasoning benchmarks. Meanwhile, studies like Mind2Web2~\citep{DBLP:journals/corr/abs-2506-21506}, GAIA2~\citep{DBLP:journals/corr/abs-2509-17158}, and RealDevWorld~\citep{DBLP:journals/corr/abs-2508-14104} demonstrate the increasing importance of agentic verifiers for agentic task evaluation. Despite this momentum, there still lacks comprehensive benchmarks to evaluate agent-as-a-judge across a wide range of agent tasks.

\subsection{Benchmarks for Task-Solving Agents}\label{sec:relatedwork:benchmark task-solving agent}

A growing body of work benchmarks task-solving LLM agents in interactive environments by evaluating end-to-end task success and trajectory quality~\citep{mohammadi2025evaluation}, covering web and application agents (e.g., WebShop~\citep{yao2022webshop}, WebArena~\citep{zhou2023webarena}, AppWorld~\citep{trivedi2024appworld}), tool-using agents that stress multi-tool planning over large APIs (e.g., ToolBench~\citep{huang2023metatool}, API-Bank~\citep{li2023api}), as well as domain-specific, robustness, and safety settings (e.g., ScienceAgentBench~\citep{chen2024scienceagentbench}, TaskBench~\citep{shen2024taskbench}). These benchmarks primarily assess an agent’s problem-solving and execution capability. TRAIL~\citep{deshpande2025trailtracereasoningagentic} further studies the evaluation of agentic systems from the perspective of trace debugging. In contrast, judge-agent benchmarks shift the focus from solving tasks to verifying behaviors, requiring agents to actively acquire evidence, inspect environment states, and audit trajectories to determine correctness.

\section{Benchmark Construction}

AJ-Bench is designed to evaluate a model’s ability to leverage external tools when verifying agent trajectories. The benchmark emphasizes three core verification dimensions: information acquisition, state verification, and process verification. To instantiate these dimensions, we curate tasks from the Search, DS, and GUI domains. Figure \ref{fig:method_figure} illustrates the overall framework of our benchmark construction and evaluation.

\begin{table*}[ht]
\centering
\begin{tabular}{l c c c c c c c c}
\toprule
\multirow{2}{*}{\textbf{Statistic}} & \multicolumn{2}{c}{\textbf{Search}} 
& \multicolumn{2}{c}{\textbf{DS}}  
& \multicolumn{3}{c}{\textbf{GUI}} & \multirow{2}{*}{\textbf{Overall}} \\
\cmidrule(lr){2-3} \cmidrule(lr){4-5} \cmidrule(lr){6-8}
 & \textbf{Wide} & \textbf{Deep} 
& \textbf{FileSystem} & \textbf{Postgres} 
& \textbf{PPT} & \textbf{Word} & \textbf{Excel} \\
\midrule
Task Count & 9 & 52 & 24 & 18 & 21 & 12 & 19 & 155 \\
Trajectory Count & 27 & 156 & 129 & 100 & 42 & 24 & 38 & 516 \\
Tool Count & 22$^\star$ & 22$^\star$ & 14 & 9 & 15$^\dagger$ & 15$^\dagger$ & 15$^\dagger$ & 60 \\
\bottomrule
\end{tabular}
\caption{Statistics of AJ-Bench across domains and subdomains, including task count, trajectory count, and tool count. $^\star$Search tasks share the same tool set. $^\dagger$GUI tasks share the same tool set.}
\label{tab:statistics}
\end{table*}

\subsection{Task Design}
\label{sec:task_design}


We design tasks by jointly considering task properties such as interactivity and complexity, and environment characteristics such as reproducibility and LLM–environment interaction for reliable evaluation.

\subsubsection{Search Domain}

We select tasks from Mind2Web2 ~\citep{DBLP:journals/corr/abs-2506-21506} and WideSearch ~\citep{wong2025widesearch}, which provide high-quality tasks with non-fixed or hard-to-exhaustively-retrieve answers and represent two complementary information-seeking paradigms. Mind2Web2 emphasizes deep search that requires multi-hop reasoning, while WideSearch focuses on wide search with broad information coverage. We exclude tasks with short, easily verifiable answers and highly time-sensitive content, such as shopping or travel, where URLs, prices, or ratings change rapidly. This filtering reflects two considerations: such tasks do not adequately test Agent-as-a-Judge capabilities, and time-sensitive tasks hinder reproducible environments and consistent evaluation.

Mind2Web2 tasks are curated through a human-in-the-loop pipeline. The details are described in the Appendix \ref{app:mind2web2_pipeline}. For WideSearch, we selected tasks that differ from those in Mind2Web2 and lightly rewrote them to explicitly encourage link-providing responses.

\subsubsection{DS Domain}

We construct the DS domain from two representative MCPMark~\citep{wu2025mcpmark} subcategories, Filesystem and Postgres, which involve manipulating file structures and database records. Task outcomes can be directly verified by inspecting the environment state, enabling reliable evaluation of the judge agent’s state-verification capability. Based on results from~\citet{wu2025mcpmark}, we exclude overly difficult tasks to maintain balance and obtain high-quality trajectories with both successes and failures, and manually remove tasks with ambiguous descriptions.


\subsubsection{GUI Domain}


We select tasks from OSWorld~\citep{xie2024osworld}, which offers a scalable real-world computer environment with high-quality multimodal agent tasks across domains. Specifically, we construct tasks from three office categories, PowerPoint, Word, and Excel, which require precise execution positions and carefully planned action sequences and thus remain challenging for current agents. We first remove tasks with unstable GUI states, such as feedback pop-up windows, and retain only those with reproducible final states under repeated execution to ensure a stable evaluation environment. We then manually review task instructions and exclude tasks with subjective elements, keeping only tasks whose completion can be verified through concrete and observable GUI state changes.

\subsection{Trajectory Collection}
\label{sec:traj_collection}

We construct trajectories for AJ-Bench using two complementary approaches: (1) leveraging existing trajectories from established benchmarks, and (2) regenerating trajectories using LLMs. All constructed trajectories are subsequently subjected to thorough human verification to ensure correctness and quality.

\subsubsection{Search Domain}

We collect trajectories from web pages using Gemini DeepResearch\footnote{\url{https://gemini.google.com/app}}, Grok DeepSearch\footnote{\url{https://grok.com/}}, and Perplexity DeepResearch\footnote{\url{https://www.perplexity.ai/}}, and manually filter out poor responses during collection. For the retained trajectories, we apply different processing strategies to Mind2Web2 and WideSearch. For Mind2Web2, we use \textit{gpt-5-2025-08-07} to extract query-relevant information from the responses, primarily to remove excessive content unrelated to the query, and apply minor human edits to fix formatting issues. For WideSearch, we perform manual extraction to preserve tables and reference links, supplemented with brief contextual explanations.

\subsubsection{DS Domain}
We construct our trajectory dataset by leveraging agent trajectories generated by multiple models on MCPMark, supplemented with original trajectories provided in the benchmark\footnote{\href{https://huggingface.co/datasets/Jakumetsu/mcpmark-trajectory-log}{https://huggingface.co/datasets/Jakumetsu/mcpmark-trajectory-log}}. To mitigate the potential bias introduced by model-specific output styles, we ensure that trajectories associated with the same task are sourced from diverse model architectures and subsequently normalized into a consistent template format. We further perform a comprehensive manual quality check to discard incomplete or noisy samples. For each task, we retain up to three successful trajectories and three failed ones, resulting in a balanced and high-quality dataset tailored for evaluating judge agents.

\subsubsection{GUI Domain}
We utilize raw action trajectories generated by multiple multimodal models from OSWorld\footnote{\href{https://huggingface.co/datasets/xlangai/ubuntu_osworld_verified_trajs}{https://huggingface.co/datasets/xlangai/ubuntu\_osworld\\\_verified\_trajs}}. To mitigate potential bias arising from differences in trajectory length, where successful trajectories typically contain fewer steps than failed ones, we deliberately select trajectories in which successful executions may involve many steps while failures terminate after relatively few steps. This strategy helps decouple task success from trajectory length. Furthermore, we account for model heterogeneity by sampling trajectories from a diverse set of models. Specifically, we extract trajectories from \textit{claude-4-sonnet-20250514-50steps}, \textit{claude-4-sonnet-20250514-15steps}, \textit{o3\_50steps}, \textit{qwen2.5-vl-32b-instruct\_100steps}, and \textit{doubao-1.5-thinking-vision-pro-250428-100step}.

\subsection{Label Annotation}
\label{sec:label_annotation}

Across all three domains, labels are binary (1/0), indicating success or failure. In the search domain, labels are defined at the item level, whereas in the other two domains, labels are assigned at the trajectory level.

\subsubsection{Search Domain}

Labels for the Mind2Web2 portion are obtained through manual annotation, where annotators assign a scoring rubric and corresponding rubric-level labels to each response. To support more fine-grained evaluation, we further employ \textit{gpt-4.1} to decompose each response into single-item units based on the rubric, enabling evaluation at a finer level of granularity. For the WideSearch portion, labels are obtained using the official WideSearch codebase. Specifically, labels are derived via majority voting across six models: \textit{gpt-4.1-2025-04-14}, \textit{gpt-5-2025-08-07}, \textit{o4-mini-2025-04-16}, \textit{claude-sonnet-4-20250514}, \textit{gemini-2.5-pro-preview-06-05}, and \textit{grok-3}. In addition, single-item units are extracted by a combination of manual refinement and rule-based parsing of the generated Markdown tables, enabling fine-grained evaluation.

\subsubsection{DS domain}
For tasks in the DS domain, a trajectory is deemed successful only if all explicit requirements in the task description are fully satisfied. We leverage the verifier scripts provided in MCPMark—derived from high-quality human annotations—to automatically determine the outcome of each trajectory, ensuring reliable supervision signals. To further guarantee label correctness, we additionally perform a manual validation pass to correct potential misjudgements and maintain consistent annotation quality.

\subsubsection{GUI Domain}

In the GUI domain, labels in OSWorld are initially assigned using rule-based scripts that compare execution trajectory outputs against golden files for office tasks. However, these scripts are inherently limited in their ability to capture all execution details and edge cases, which may lead to mislabeling. To ensure label reliability, we therefore manually inspect each trajectory to verify its correctness.

\subsection{Dataset Statistics}

\begin{figure}[t] 
    \centering
    \includegraphics[width=1.0\linewidth]{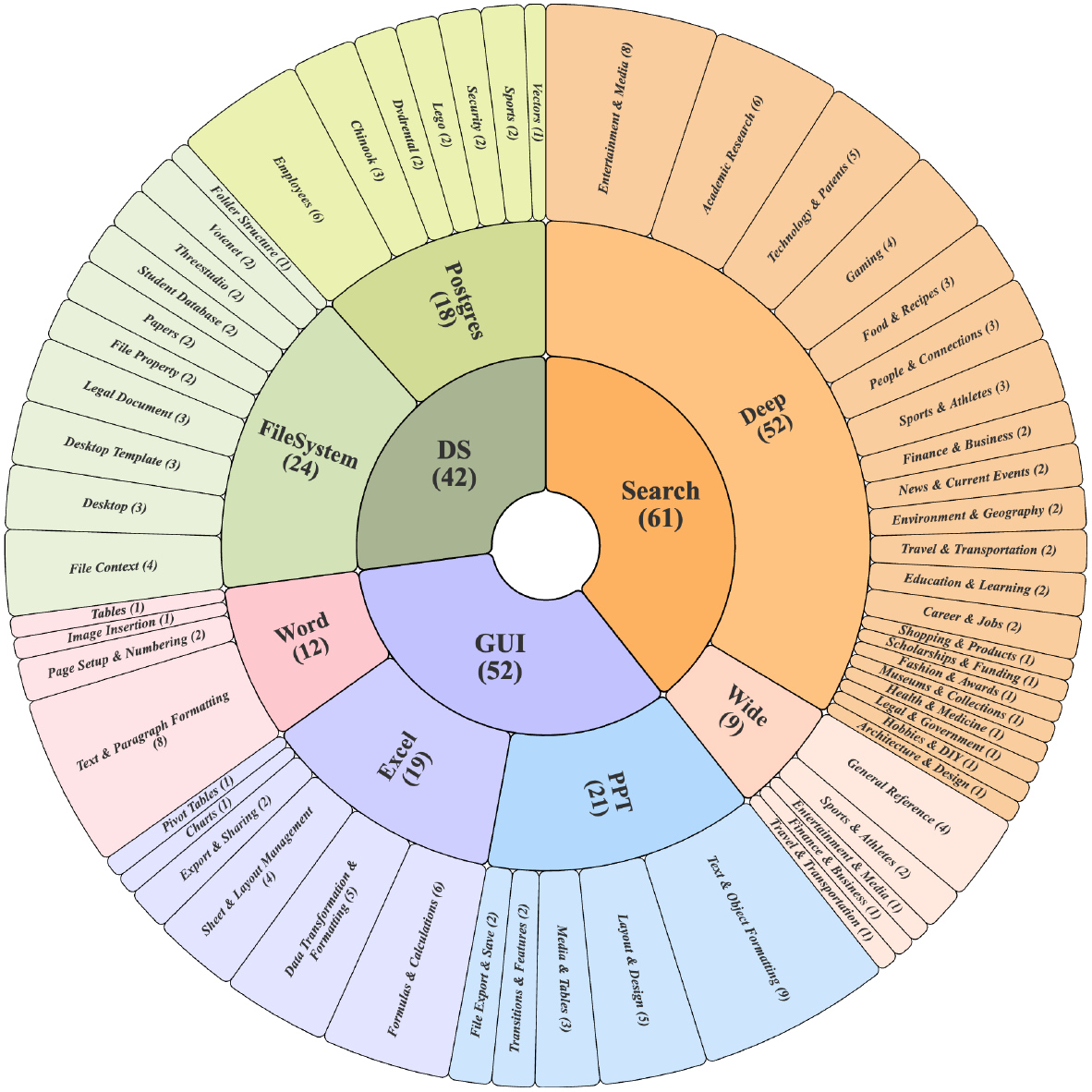} 
    \caption{Task distribution of AJ-Bench}
    \label{fig:distribution_figure}
    \vspace{-10pt}
\end{figure}

Following the construction pipeline described in \S\ref{sec:task_design}–\S\ref{sec:label_annotation}, we obtain a total of 155 tasks, 516 trajectories, and 60 different tools. Detailed statistics for each domain are reported in Table \ref{tab:statistics}. In addition, we analyze the distribution of task types across different domains, as illustrated in Figure \ref{fig:distribution_figure}.

\subsection{Environment Construction}

Our benchmark provides raw action trajectories and task-specific configurations for both the DS and GUI domains. To support interactive evaluation, we replay the final environment state of each task, enabling the agent to interact with a live environment and actively acquire additional contextual information beyond static trajectories.

In both the DS and GUI domains, evaluation trajectories are replayed by sequentially executing the extracted action sequences in trajectories to reconstruct independent environments that support concurrent evaluation. DS tasks are replayed locally, while GUI tasks are deployed on isolated AWS instances. Once the environment is restored to its final recorded state, the agent begins to evaluate through interaction.



\section{Experiments}

\begin{table*}[t]
\centering
\scalebox{0.8}{
\begin{tabular}{lccccccccc}
\toprule
\multirow{2}{*}{\textbf{Model}} & \multirow{2}{*}{\textbf{Agentic}}
& \multicolumn{2}{c}{\textbf{Search}} 
& \multicolumn{2}{c}{\textbf{DS}}  
& \multicolumn{3}{c}{\textbf{GUI}}
& \multirow{2}{*}{\makecell{\textbf{Overall}\\\textbf{Avg@3}}} \\
\cmidrule(lr){3-4} \cmidrule(lr){5-6} \cmidrule(lr){7-9}
&
& \textbf{Wide} 
& \textbf{Deep} 
& \textbf{FileSystem} 
& \textbf{Postgres} 
& \textbf{PPT} 
& \textbf{Word} 
& \textbf{Excel}
& \\
\midrule


\multicolumn{10}{l}{\textbf{Proprietary Models}} \\
\midrule

\modelicon{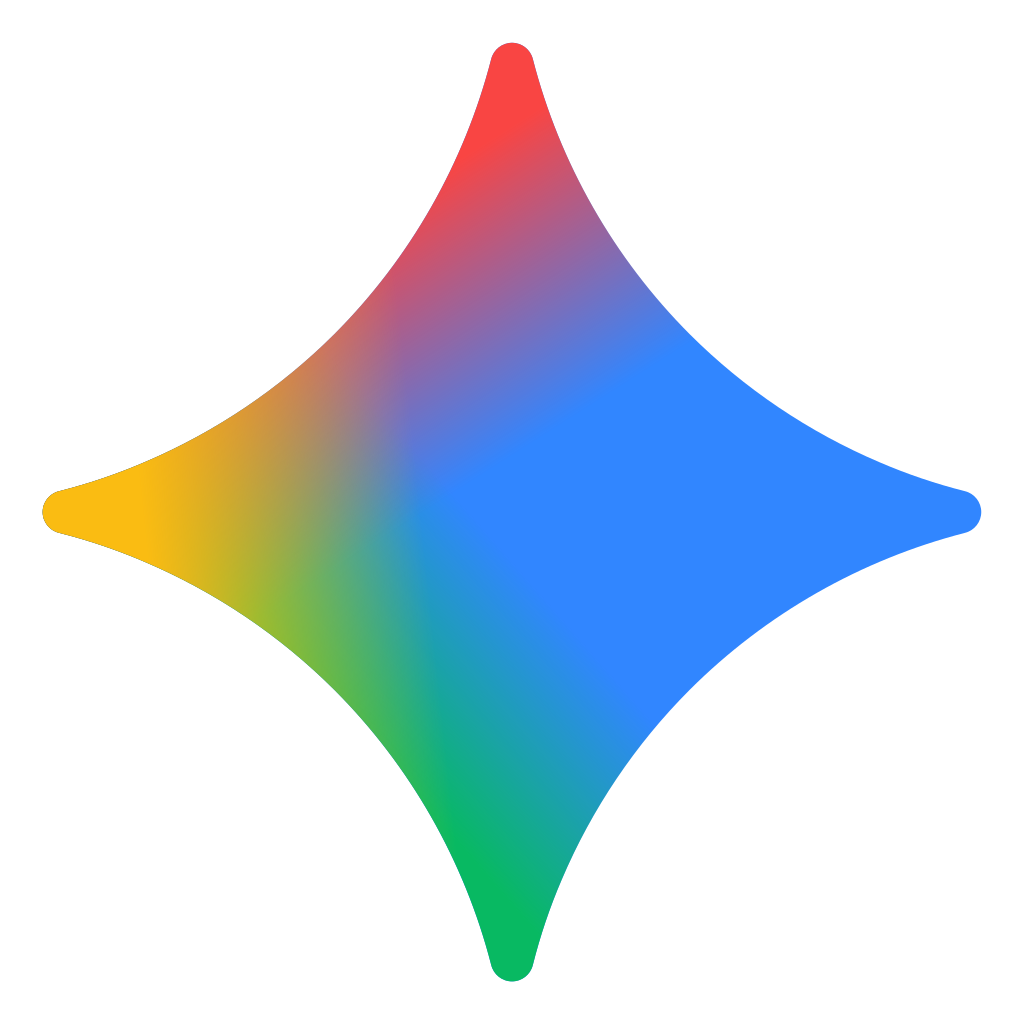}{gemini-3-pro-preview} & \xmark
& 72.70 & 81.26
& 75.69 & 73.20
& 76.10 & 72.14 & 74.28
& 75.05{\scriptsize\textcolor{blue!70!black}{$\pm$1.26}}\\

\modelicon{gemini.png}{gemini-2.5-pro} & \xmark
& 66.35 & 81.22
& 66.10 & 68.96
& 68.72 & 60.13 & 66.67
& 68.31{\scriptsize\textcolor{blue!70!black}{$\pm$0.95}}\\

\modelicon{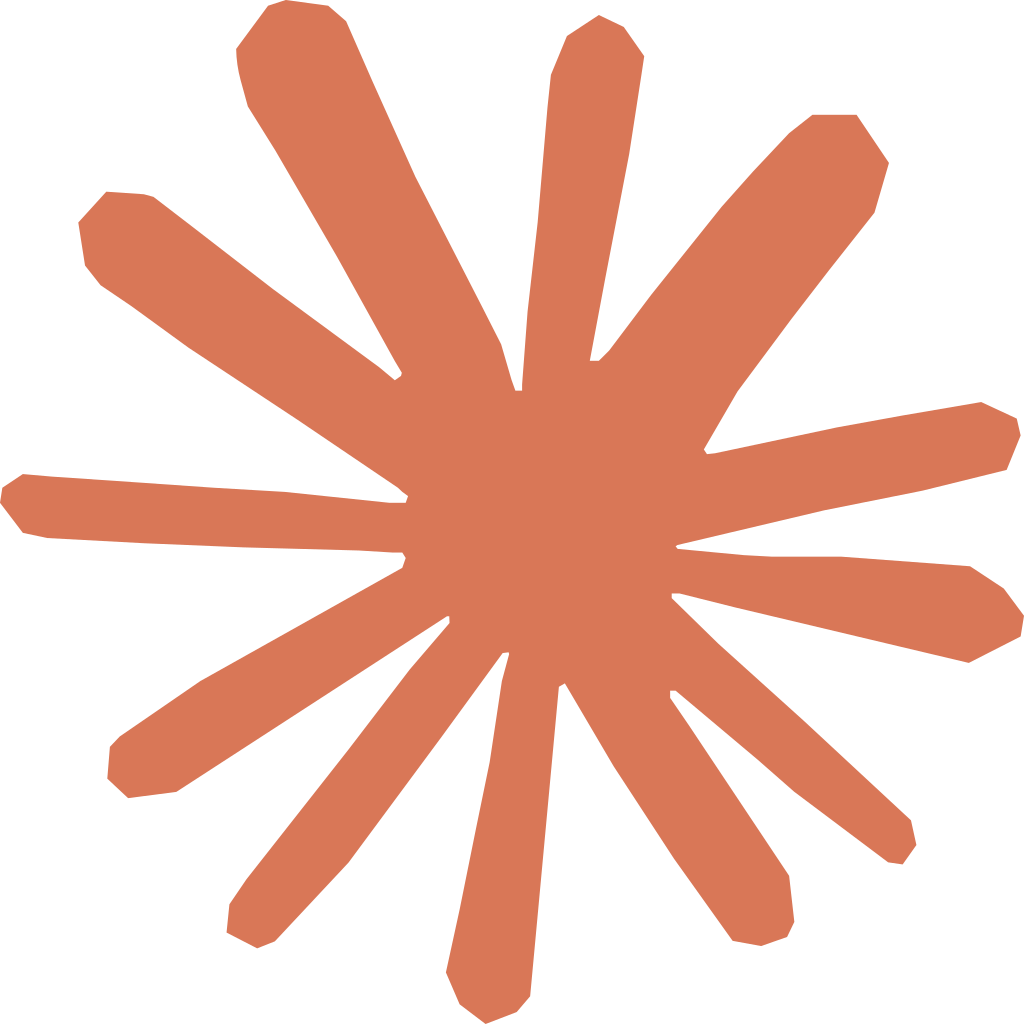}{claude-opus-4.5}  & \xmark
& 64.26 & 81.11
& 66.06 & 69.66
& 59.21 & 51.45 & 75.77
& 66.79{\scriptsize\textcolor{blue!70!black}{$\pm$1.33}}\\

\modelicon{claude.png}{claude-sonnet-4.5} & \xmark
& 61.02 & 81.34
& 69.26 & 68.36
& 75.61 & 61.56 & 71.24
& 69.77{\scriptsize\textcolor{blue!70!black}{$\pm$1.18}}\\

\modelicon{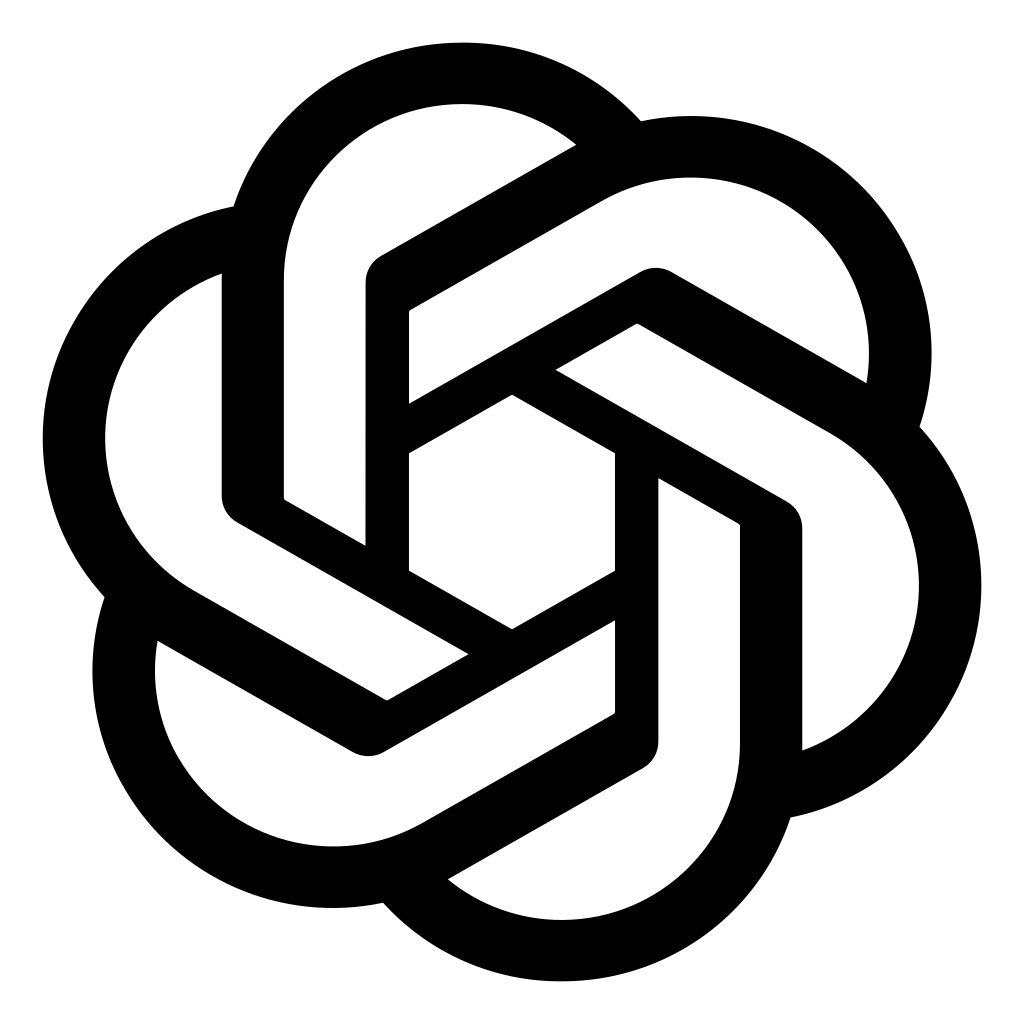}{gpt-5} & \xmark
& 66.33 & 80.37
& 59.09 & 62.84
& 51.90 & 44.81 & 61.78
& 61.02{\scriptsize\textcolor{blue!70!black}{$\pm$0.13}}\\

\modelicon{gpt.png}{gpt-5.1} & \xmark
& 58.02 & 70.90
& 46.27 & 57.53
& 41.90 & 39.54 & 60.33
& 53.50{\scriptsize\textcolor{blue!70!black}{$\pm$3.56}}\\

\modelicon{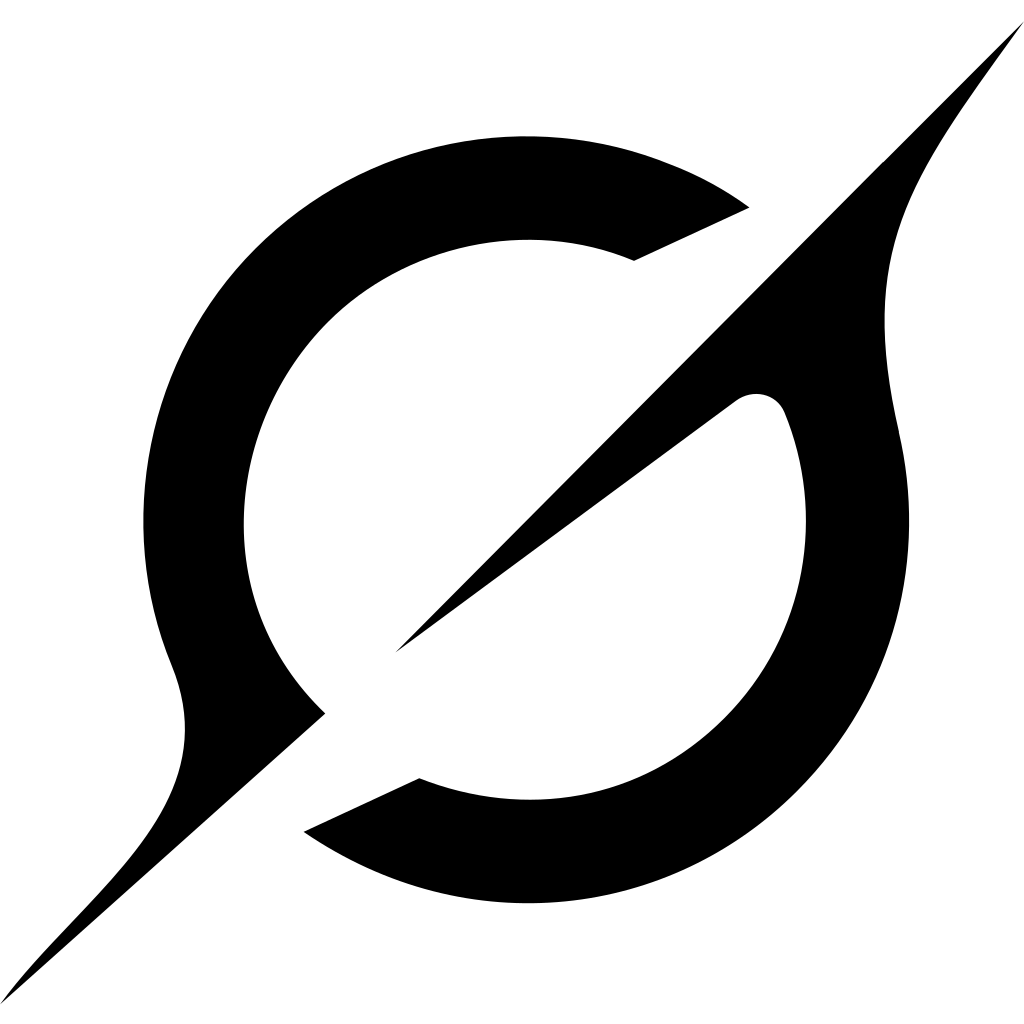}{grok-4}  & \xmark
& 69.18 & 78.32
& 75.70 & 59.57
& 61.11 & 65.26 & 75.52
& 69.24{\scriptsize\textcolor{blue!70!black}{$\pm$1.11}}\\

\midrule

\modelicon{gpt.png}{gpt-5-mini-low} & \xmark
& 60.84 & 68.42
& 60.41 & 65.52
& 45.05 & 48.41 & 64.36
& 59.00{\scriptsize\textcolor{blue!70!black}{$\pm$0.91}}\\

\rowcolor{gray!20} \modelicon{gpt.png}{gpt-5-mini-low} & \cmark
& 65.93 & 75.69
& 67.54 & 67.30
& 76.28 & 72.22 & 81.89
& 72.41{\scriptsize\textcolor{blue!70!black}{$\pm$1.68}}\\

\rowcolor{gray!60}\modelicon{gpt.png}{\textbf{Improvement}} & \cmark
& \textbf{+5.09} & \textbf{+7.27}
& \textbf{+7.13} & \textbf{+1.78}
& \textbf{+31.23} & \textbf{+23.81} & \textbf{+17.53}
& \textbf{+13.41}\\

\midrule

\multicolumn{9}{l}{\textbf{Open-Source Models}} \\
\midrule

\modelicon{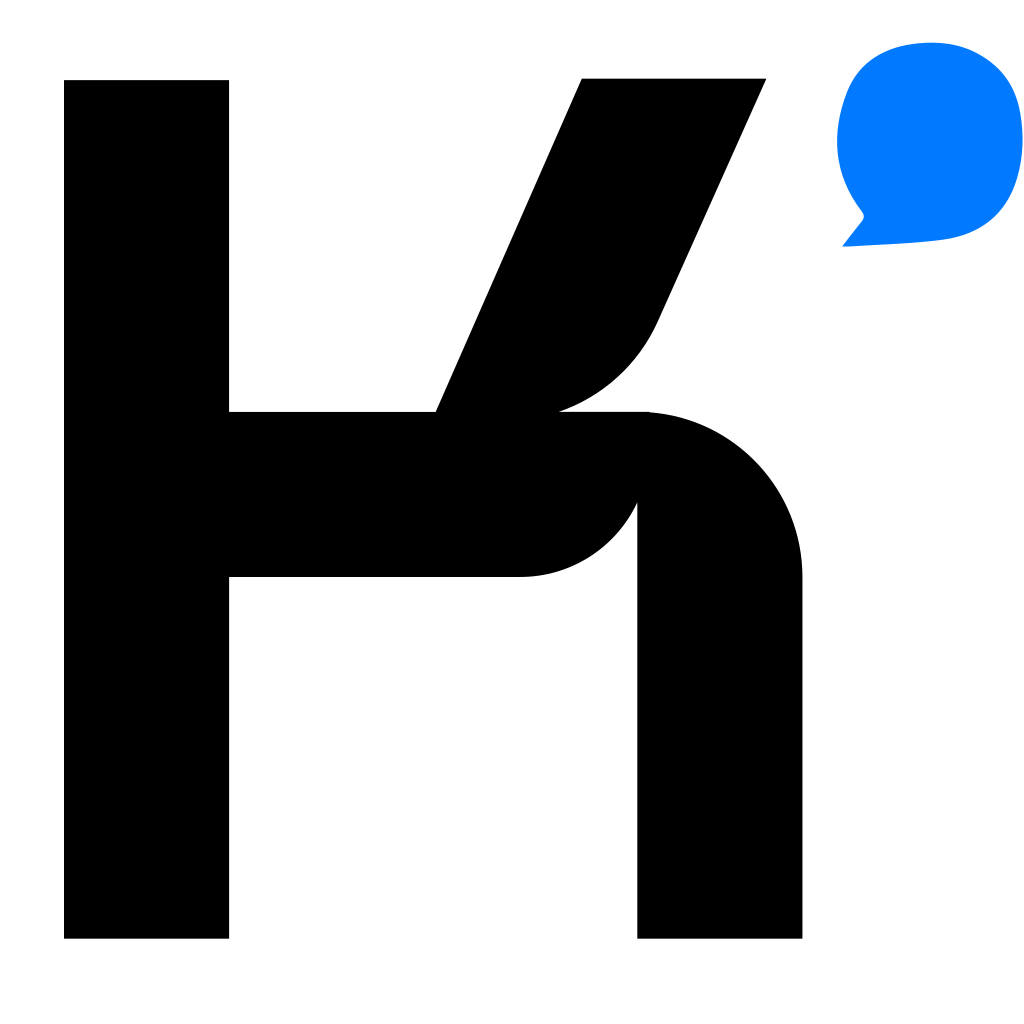}{kimi-k2-0905-preview} & \xmark
& 63.52 & 80.17
& 55.96 & 65.85
& 65.53 & 55.39 & 63.90
& 64.33{\scriptsize\textcolor{blue!70!black}{$\pm$2.07}}\\

\modelicon{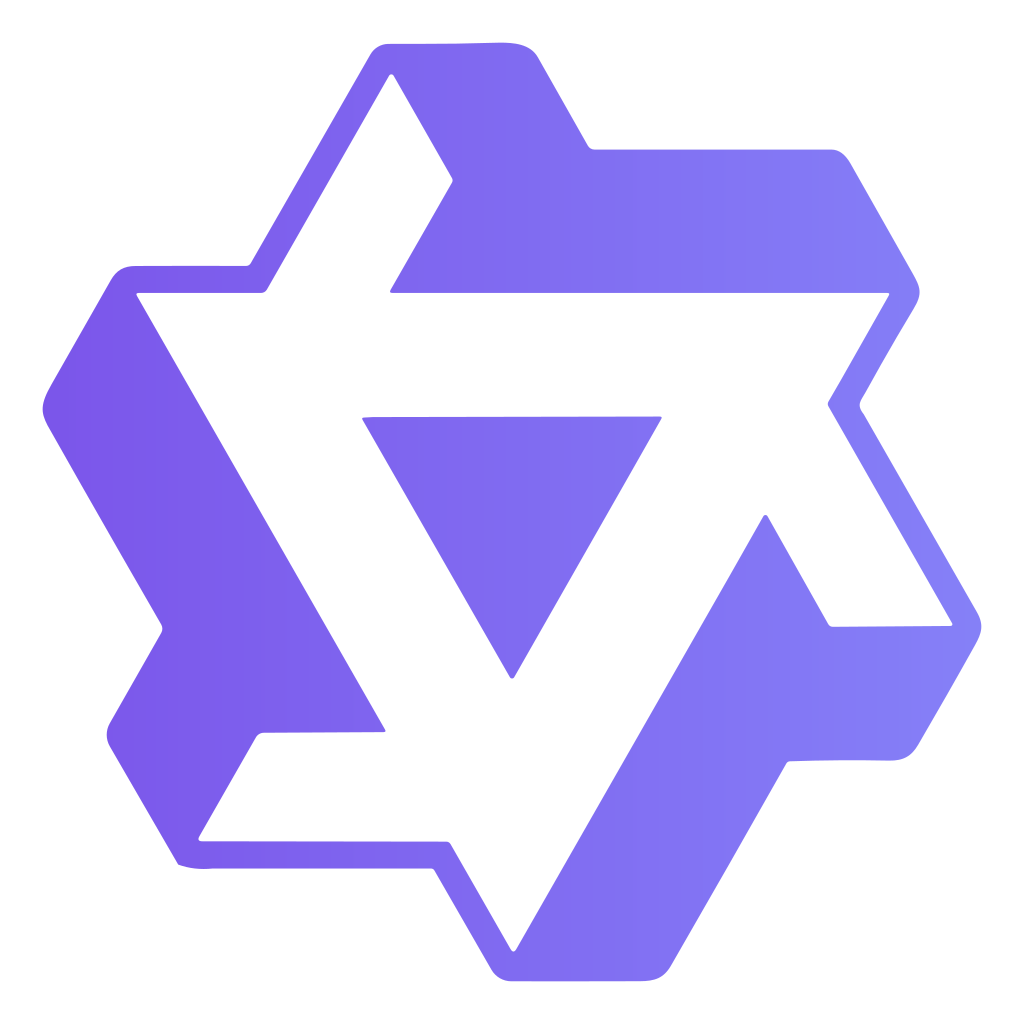}{qwen3-235b-a22b} & \xmark
& 62.69 & 81.33
& 64.66 & 64.32
& 45.50 & 36.82 & 53.97
& 58.47{\scriptsize\textcolor{blue!70!black}{$\pm$2.32}}\\

\modelicon{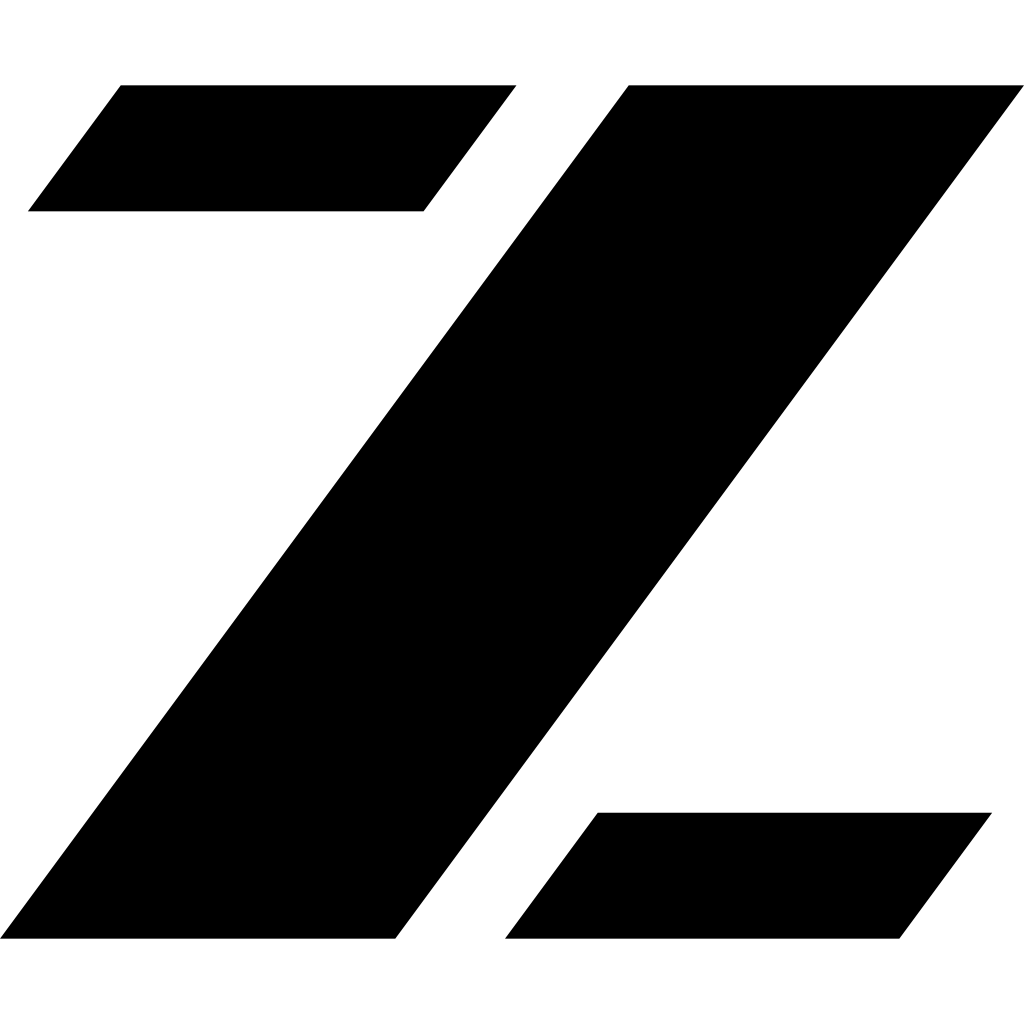}{glm-4.6} & \xmark
& 66.61 & 77.88
& 60.86 & 64.94
& 60.82 & 50.07 & 72.49
& 64.81{\scriptsize\textcolor{blue!70!black}{$\pm$0.96}}\\

\modelicon{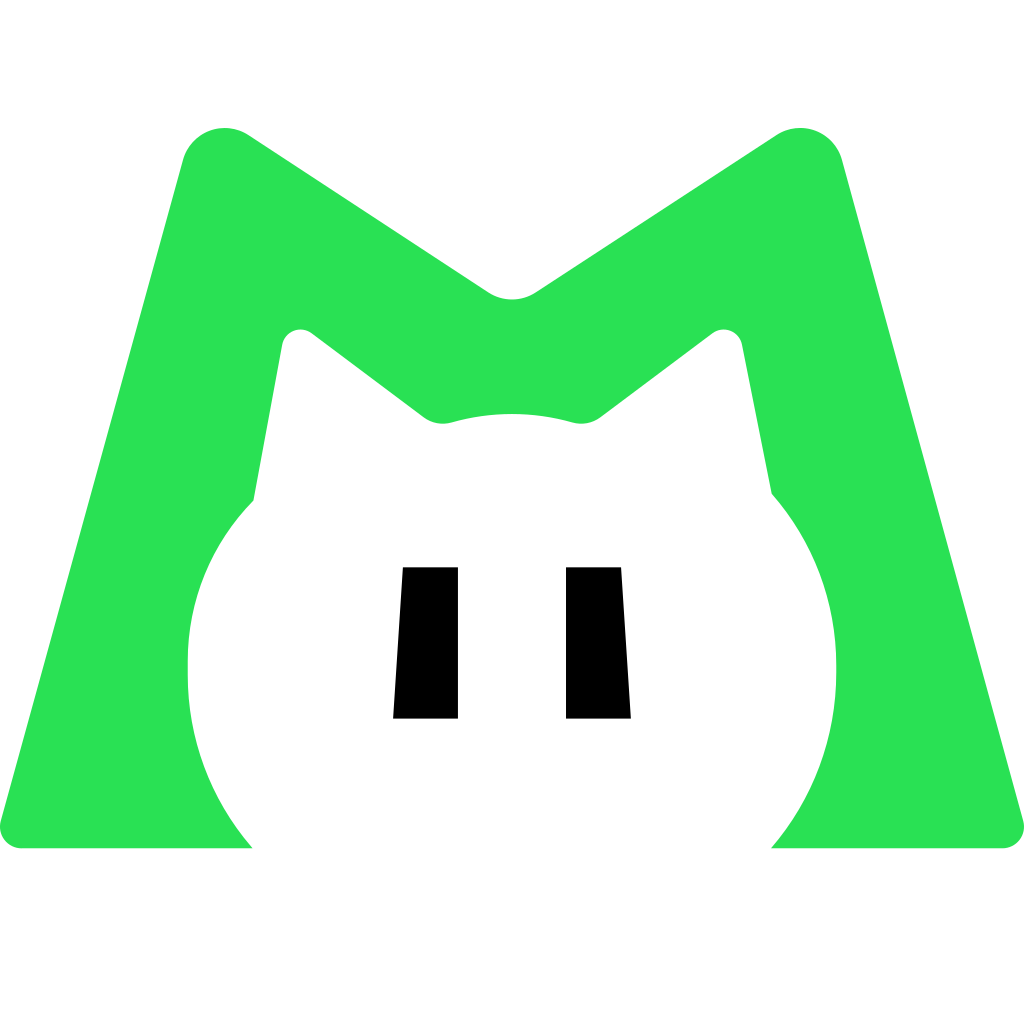}{longcat-flash-chat} & \xmark
& 64.44 & 81.80
& 59.13 & 65.54
& 45.33 & 30.35 & 55.88
& 57.50{\scriptsize\textcolor{blue!70!black}{$\pm$3.19}}\\

\midrule

\modelicon{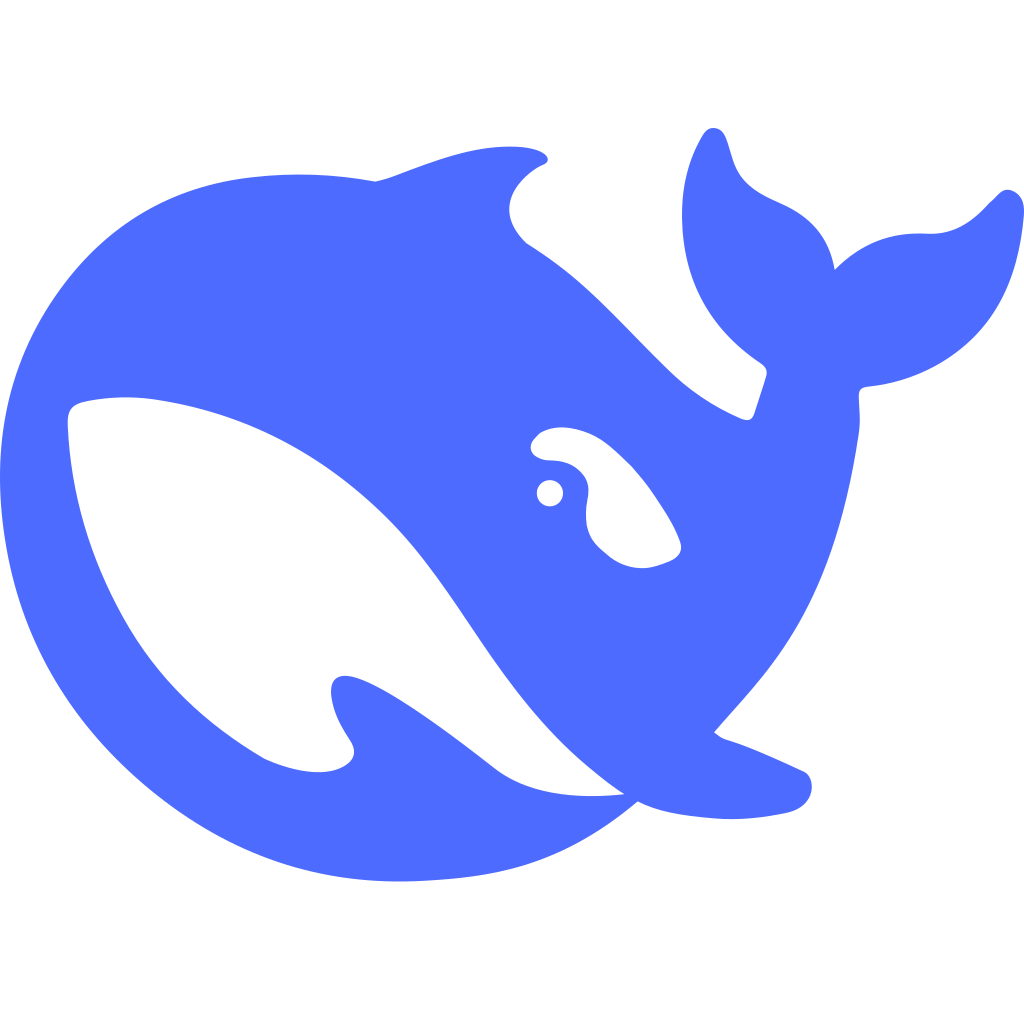}{deepseek-v3.2} & \xmark
& 63.65 & 62.91
& 60.31 & 66.31
& 58.38 & 69.77 & 70.12
& 64.49{\scriptsize\textcolor{blue!70!black}{$\pm$0.50}}\\

\rowcolor{gray!20} \modelicon{deepseek.png}{deepseek-v3.2}  & \cmark   
& 72.47 & 82.14
& 72.60 & 72.70
& 83.14 & 78.64 & 79.71
& 77.34{\scriptsize\textcolor{blue!70!black}{$\pm$1.36}}\\

\rowcolor{gray!60}\modelicon{deepseek.png}{\textbf{Improvement}} & \cmark
& \textbf{+8.82} & \textbf{+19.23}
& \textbf{+12.29} & \textbf{+6.39}
& \textbf{+24.76} & \textbf{+8.87} & \textbf{+9.59}
& \textbf{+12.85}\\

\bottomrule
\end{tabular}
}

\caption{Performance Comparison under LLM-as-a-Judge and Agent-as-a-Judge Settings}
\label{tab:judge_comparison}
\end{table*}

In this section, we first describe the experimental setup (\S \ref{sec:exp_setup}), then demonstrate the superiority of Agent-as-a-Judge over LLM-as-a-Judge (\S \ref{sec:exp_main}), and finally conduct ablation studies to investigate the factors influencing Agent-as-a-Judge performance from the perspectives of internal capabilities and external information inputs (\S~\ref{sec:exp_ablation}).

\subsection{Setup}
\label{sec:exp_setup}




\paragraph{Models.} For LLM-as-a-Judge, we evaluate a range of sota closed-source models, including the Gemini family (i.e., \textit{gemini-3-pro-preview} ~\citep{gemini3blog} and \textit{gemini-2.5-pro} ~\citep{comanici2025gemini}), the Claude family (i.e., \textit{claude-opus-4.5} and \textit{claude-sonnet-4.5} ~\citep{claude_opus_4_5, claude_sonnet_4_5}), the GPT family (i.e., \textit{gpt-5}, \textit{gpt-5-mini}, and \textit{gpt-5.1} ~\citep{gpt_5, gpt_5_1}), as well as \textit{grok-4} ~\citep{grok_4}. We also consider several strong open-source models: \textit{kimi-k2-0905-preview} ~\citep{team2025kimi}, \textit{qwen3-235b-a22b} ~\citep{yang2025qwen3}, \textit{glm-4.6} ~\citep{glm_4_6},  \textit{longcat-flash-chat} ~\citep{team2025longcat}, and \textit{deepseek-v3.2} ~\citep{liu2025deepseek}. For Agent-as-a-Judge, due to budget constraints, we select \textit{gpt-5-mini-low} as the representative closed-source model and \textit{deepseek-v3.2} as the representative open-source model.


\paragraph{Implementation Details.} Our agent implementation is built on MCPMark\footnote{\url{https://github.com/eval-sys/mcpmark}}, a framework designed to evaluate an LLM’s intrinsic ability to decide when and how to invoke tools, without relying on complex or heavily engineered workflows. Unless explicitly stated otherwise, all models are evaluated with their default configurations (e.g., temperature and reasoning effort). We adopt F1 score as our primary evaluation metric. For the metric design, in the Search domain, we aggregate the evaluations of all single items within a trajectory into a single result, from which the F1 score is computed. In the DS and GUI domains, we compute the F1 score based on trajectory-level evaluations. Results reported in Table~\ref{tab:judge_comparison} are averaged over three runs. Further details are available in the Appendix \ref{app:implementation_details}.

\subsection{Main Experiments}
\label{sec:exp_main}

We evaluate a range of open-source and closed-source models on AJ-Bench, with the results summarized in Table \ref{tab:judge_comparison}. These results lead to the following observations:

\paragraph{Agent-as-a-Judge consistently outperforms LLM-as-a-Judge when built upon the same base model.} For a given model—whether a thinking model or a chat model—enabling tool usage leads to an average improvement of approximately 13 percentage points across the three domains compared to its counterpart without tool calls. These results demonstrate the substantial potential of Agent-as-a-Judge, highlighting its effectiveness as a verifier for more reliable trajectory evaluation.

\paragraph{Agent-as-a-Judge built on weaker models can achieve performance comparable to that of LLM-as-a-Judge based on sota models.} Our implemented Agent-as-a-Judge baseline consistently outperforms existing sota LLM-as-a-Judge models. This result highlights the inherent limitations of LLM-as-a-Judge. For tasks that require effective interaction with the external environment, Agent-as-a-Judge demonstrates clear advantages over LLM-as-a-Judge.

\subsection{Ablation Study}
\label{sec:exp_ablation}

\begin{table*}[t]

\centering
\renewcommand{\arraystretch}{1.1} 
\scalebox{0.85}{
\begin{tabular}{l c c c c c c c c c}
\toprule
\multirow{2}{*}{\textbf{Model}} & \multirow{2}{*}{\textbf{Reasoning}} & \multicolumn{2}{c}{\textbf{Search}} & \multicolumn{2}{c}{\textbf{DS}} & \multicolumn{3}{c}{\textbf{GUI}} & \multirow{2}{*}{\textbf{Overall}} \\
\cmidrule(lr){3-4} \cmidrule(lr){5-6} \cmidrule(lr){7-9}
& & \textbf{Wide} & \textbf{Deep} & \textbf{FileSystem} & \textbf{Postgres} & \textbf{PPT} & \textbf{Word} & \textbf{Excel} & \\
\midrule
\multirow{3}{*}{\modelicon{gpt.png}{gpt-5-mini}}
& low & 65.93 & 75.69
& 67.54 & 67.30
& 76.28 & 72.22 & 81.89
& 72.41\\
& medium & 72.76 & 77.11
& 75.80 &  69.84
& 82.05 & 72.00 & 82.35
& 75.99\\
& high & 74.48 & 79.19
& 71.53 &  67.92
& 78.95 & 63.64 & 81.08
& 73.83\\
\midrule
\multirow{2}{*}{\modelicon{deepseek.png}{deepseek-v3.2}}
& N/A & 72.47 & 82.14
& 72.60 & 72.70
& 83.14 & 78.64 & 79.71
& 77.34\\
& thinking & 70.37 & 79.31
& 68.83 &  74.13
& 82.05 & 78.57 & 86.49
& 77.11\\
\bottomrule
\end{tabular}
}
\caption{Performance comparison of models under different reasoning effort settings across tasks}
\label{tab:thinking_ablation}
\end{table*}

To investigate the effectiveness of Agent-as-a-Judge, we conduct ablation studies from two perspectives: the model’s internal capabilities and external information inputs. From the perspective of internal capabilities, we examine the impact of reasoning ability on the performance of Agent-as-a-Judge (\S \ref{sec:thinking_ablation}). From the perspective of external information inputs, we study the effects of tool invocation frequency and input modality on Agent-as-a-Judge performance (\S \ref{sec:turns_ablation}–\S \ref{sec:multimodal_ablation}).

\subsubsection{Thinking Ablation}
\label{sec:thinking_ablation}

In this section, we investigate the impact of different levels of reasoning effort on Agent-as-a-Judge when using the same base model. As shown in Table \ref{tab:thinking_ablation}, across all three domains, we observe that for \textit{gpt-5-mini}, the \texttt{medium} setting generally outperforms the \texttt{low} setting, while the \texttt{high} setting does not consistently outperform \texttt{medium}. For \textit{deepseek-v3.2}, the \texttt{thinking} variant performs worse than the \texttt{no-thinking} variant. These results indicate that increasing reasoning effort does not necessarily enhance the performance of Agent-as-a-Judge. In other words, stronger intrinsic reasoning capability is not equivalent to the ability to effectively invoke tools, analyze tool outputs, and make reliable decisions.


\subsubsection{Interaction Turns Ablation}
\label{sec:turns_ablation}

\begin{figure}[t] 
    \centering
    \includegraphics[width=1.0\linewidth]{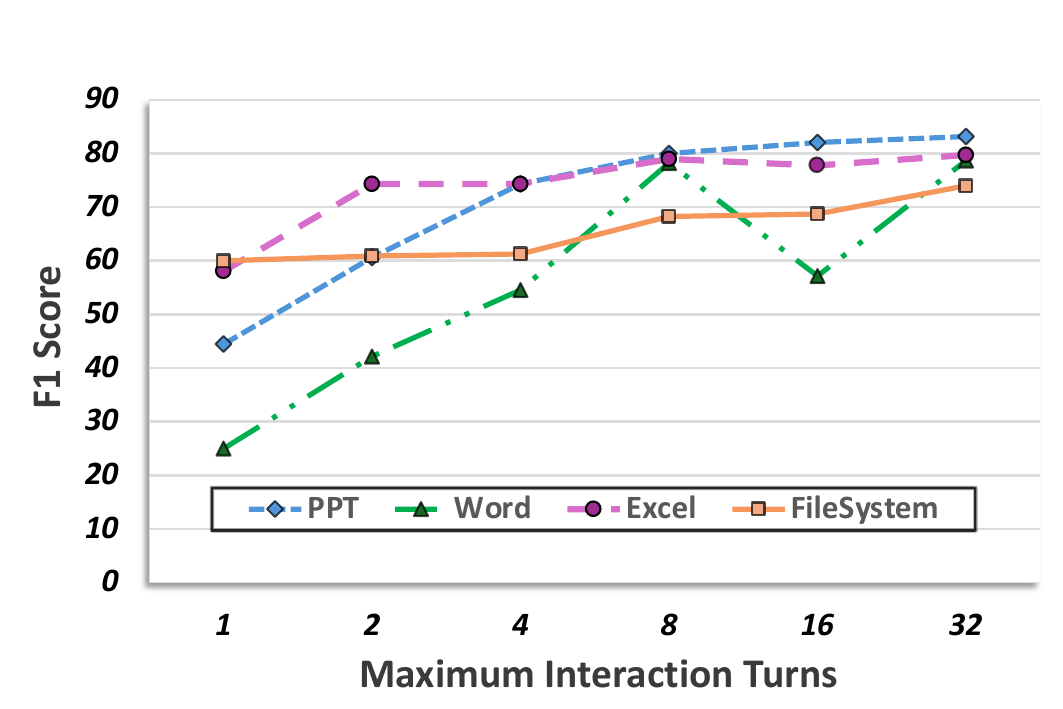}
    \caption{Comparing different interaction turns' effect on the evaluation results}
    \label{fig:interaction_turns_ablation}
\end{figure}

To investigate the effect of the interaction budget on the agent's evaluation ability, we set different maximum interaction turn limits under \textit{deepseek-v3.2} for several subdomains. As shown in Figure~\ref{fig:interaction_turns_ablation}, increasing the interaction budget consistently improves F1 scores across all tasks, with the most pronounced gains observed for smaller budgets. This suggests that more information is retrieved through interaction with the environment, which notably aids evaluation. Furthermore, task domains vary in their sensitivity to the interaction budget. Word and PPT tasks benefit more from extended interactions, indicating a greater reliance on iterative information gathering.


\subsubsection{Multimodal Ablation}
\label{sec:multimodal_ablation}
\begin{figure}[t] 
    \centering
    \includegraphics[width=1.0\linewidth]{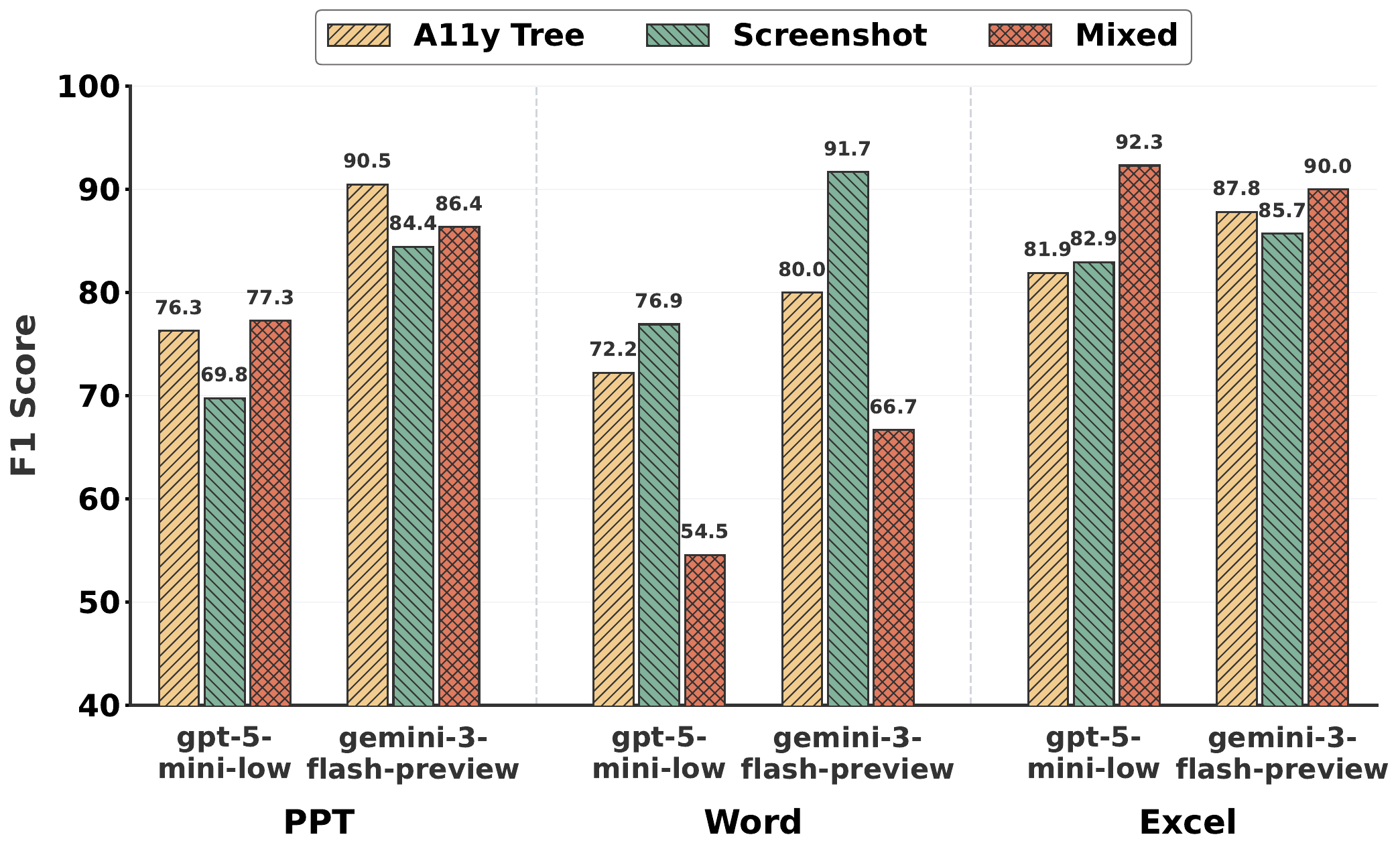} 
    \caption{Comparing the effect of each modality on agent performance under two models (\textit{gpt-5-mini-low} and \textit{gemini-3-flash-preview}) on the GUI domain}
    \label{fig:multimodal_ablation}
\end{figure}

We conduct a modality ablation study in the GUI domain to examine the impact of multimodal inputs on evaluation performance. The agent leverages information from the live environment, including the accessibility tree and screenshots, under three input configurations: (i) accessibility tree only, (ii) screenshot only, and (iii) a combination of both. The study uses two multimodal models, \textit{gpt-5-mini-low} and \textit{gemini-3-flash-preview}.
As shown in Figure~\ref{fig:multimodal_ablation}, the effectiveness of different multimodal modalities varies substantially across subdomains. In the PPT subdomain, the accessibility tree and the mixed setting achieve comparable performance. For Word tasks, the screenshot modality yields the best results. And in the Excel subdomain, the mixed-modality configuration consistently outperforms the single-modality alternatives. These findings suggest that incorporating multiple modalities does not uniformly improve agent performance. Instead, mixed inputs can introduce noise and redundancy that may distract the agent and degrade decision-making in certain scenarios.

\section{Conclusions}

In this work, we identify a critical gap in existing benchmarks, namely the lack of systematic evaluation for Agent-as-a-Judge, and introduce AJ-Bench as the first benchmark specifically designed for this purpose. AJ-Bench covers three domains, Search, DS, and GUI, comprising 155 tasks and 516 trajectories for comprehensive Agent-as-a-Judge evaluation. Our simple yet effective Agent-as-a-Judge baseline demonstrates strong performance and consistently outperforms LLM-as-a-Judge, highlighting the promise of agent-based judging paradigms. We hope that AJ-Bench will serve as a foundational evaluation platform and a valuable resource for the community, facilitating future research on Agent-as-a-Judge.

\section*{Limitations}

\paragraph{Task Diversity and Scalability.} Most tasks in AJ-Bench are adapted from existing benchmarks through modification rather than being created entirely from scratch. In future work, we plan to construct a larger portion of tasks independently and to scale up the data generation pipeline, enabling broader coverage and potential use in training settings.

\paragraph{Environment Stability.} In the search domain, Agent-as-a-Judge relies on interactions with external web environments. As a result, instability in network connectivity may affect evaluation reliability.

\section{Acknowledgement}

This research was supported by Meituan.

\bibliography{custom}

\clearpage

\appendix

\label{sec:appendix}

\begin{table*}[t]
\centering
\begin{tabular}{l l c c c c c c}
\toprule
\multirow{2}{*}{\textbf{Framework}} & \multirow{2}{*}{\textbf{Model}} & \multirow{2}{*}{\textbf{Agentic}} & \multicolumn{2}{c}{\textbf{Search}} & \multicolumn{3}{c}{\textbf{GUI}} \\
\cmidrule(lr){4-5} \cmidrule(lr){6-8}
& & & \textbf{Wide} & \textbf{Deep} & \textbf{PPT} & \textbf{Word} & \textbf{Excel} \\
\midrule

LLM-Judge & gpt-5-mini-low & \xmark & 60.84 & 68.42 & 45.05 & 48.41 & 64.36 \\
MCPMark   & gpt-5-mini-low & \cmark & 65.93 & 75.69 & 76.28 & 72.22 & 81.89 \\
ReAct     & gpt-5-mini-low & \cmark & 51.13 & 71.84 & 64.86 & 63.64 & 76.47 \\
\midrule

LLM-Judge & deepseek-v3.2 & \xmark & 63.65 & 62.91 & 58.38 & 69.77 & 70.12 \\
MCPMark   & deepseek-v3.2 & \cmark & 72.47 & 82.14 & 83.14 & 78.64 & 79.71 \\
ReAct     & deepseek-v3.2 & \cmark & 70.51 & 77.88 & 95.24 & 75.00 & 85.17 \\
\bottomrule
\end{tabular}
\caption{Ablation study of agent frameworks across two backbone models.}
\label{tab:framework_ablation}
\end{table*}

\begin{table*}[t]
\centering
\begin{tabular}{l c c c c}
\toprule
\textbf{Model} & \textbf{Agentic} & \textbf{Wide} & \textbf{FileSystem} & \textbf{PPT} \\
\midrule
gemini-3-flash-preview & $\times$ & 69.13 & 76.54 & 78.26 \\
gemini-3-flash-preview & $\checkmark$ & \textbf{72.94} & \textbf{78.62} & \textbf{90.48} \\
\midrule
claude-sonnet-4.5 & $\times$ & 61.02 & \textbf{69.76} & 75.61 \\
claude-sonnet-4.5 & $\checkmark$ & \textbf{69.87} & 62.72 & \textbf{78.26} \\
\midrule
kimi-k2.5 & $\times$ & 67.40 & 58.18 & 69.77 \\
kimi-k2.5 & $\checkmark$ & \textbf{72.09} & \textbf{62.03} & \textbf{80.00} \\
\midrule
glm-4.7 & $\times$ & 64.20 & 59.65 & 40.00 \\
glm-4.7 & $\checkmark$ & \textbf{71.96} & \textbf{63.16} & \textbf{80.00} \\
\bottomrule
\end{tabular}
\caption{Performance comparison of additional judge models with and without the Agent-as-a-Judge setting on a representative subset of tasks. We report results on three subdomains: Wide, FileSystem, and PPT.}
\label{tab:appendix_agentic_judge_models}
\end{table*}

\begin{table*}[t]
\centering
\small
\setlength{\tabcolsep}{3pt}
\begin{tabular}{lc*{7}{cc}}
\toprule
\multirow{2}{*}{Model} & \multirow{2}{*}{Agentic}
& \multicolumn{2}{c}{Wide}
& \multicolumn{2}{c}{Deep}
& \multicolumn{2}{c}{FileSystem}
& \multicolumn{2}{c}{Postgres}
& \multicolumn{2}{c}{PPT}
& \multicolumn{2}{c}{Word}
& \multicolumn{2}{c}{Excel} \\
\cmidrule(lr){3-4}\cmidrule(lr){5-6}\cmidrule(lr){7-8}\cmidrule(lr){9-10}
\cmidrule(lr){11-12}\cmidrule(lr){13-14}\cmidrule(lr){15-16}
& & L & U & L & U & L & U & L & U & L & U & L & U & L & U \\
\midrule
gpt-5-mini-low & $\times$ & 60.68 & 61.00 & 66.18 & 70.66 & 58.00 & 62.82 & 64.03 & 67.01 & 32.63 & 57.47 & 39.89 & 56.93 & 57.90 & 70.82 \\
gpt-5-mini-low & $\checkmark$ & 59.90 & 71.96 & 74.89 & 76.49 & 60.36 & 74.72 & 62.15 & 72.45 & 73.53 & 79.03 & 60.27 & 84.17 & 77.64 & 86.14 \\
deepseek-v3.2 & $\times$ & 62.76 & 64.54 & 61.19 & 64.63 & 50.76 & 69.86 & 62.33 & 70.29 & 42.08 & 74.68 & 44.90 & 94.64 & 65.76 & 74.48 \\
deepseek-v3.2 & $\checkmark$ & 70.95 & 73.99 & 80.99 & 83.29 & 71.49 & 73.71 & 70.45 & 74.95 & 67.30 & 98.98 & 63.87 & 93.41 & 63.72 & 95.70 \\
\bottomrule
\end{tabular}
\caption{95\% confidence intervals of subdomain-level performance over three independent runs.}
\label{tab:subdomain_ci}
\end{table*}

\section{Appendix}

\subsection{Mind2Web2 Task Design}
\label{app:mind2web2_pipeline}

Tasks are first categorized into three groups: \texttt{ground\_truth}, \texttt{no\_ground\_truth}, and \texttt{time\_sensitive}. \texttt{Ground\_truth} tasks have fixed, well-defined answers, whereas \texttt{no\_ground\_truth} tasks do not admit a single correct answer—for example, tasks where multiple valid solutions exist and the task only requires returning a subset (e.g., three or five items). \texttt{Time\_sensitive} tasks are filtered or rewritten into the other two categories, followed by further filtering of \texttt{ground\_truth} tasks, some of which are rewritten as \texttt{no\_ground\_truth}. Finally, \texttt{no\_ground\_truth} tasks are re-checked for potential time sensitivity or implicit ground truth and rewritten if needed. The resulting Mind2Web2 subset contains only \texttt{no\_ground\_truth} tasks.

\subsection{Human Annotation}

We employ a dedicated data annotation team to label Mind2Web2 data in the Search domain. The team consists of full-time annotators and student annotators, whose compensation is comparable to local market rates for similar roles. Prior to annotation, we provide several representative Mind2Web2 examples as references. Annotators are required to first formulate evaluation rubrics for each response and then assign labels for every criterion in the rubric.

\subsection{Implementation Details}
\label{app:implementation_details}

\subsubsection{Search Domain}

Because the pages returned in the Search domain are often lengthy, we apply summarization using the same model as the agent to extract the page content, and use the resulting summary as the agent’s context. Specifically, for \textit{gpt-5-mini}, we use the low reasoning effort configuration during summarization, whereas for \textit{deepseek-v3.2}, explicit reasoning is disabled (no thinking) for the summarization stage.

\subsubsection{GUI Domain}
In the GUI domain, we constructed our evaluation pipeline using OSWorld components within the MCPMark framework. Specifically, we implemented an OSWorld MCP server and MCP client to integrate with MCPMark. To enable highly parallelized evaluation, we leverage AWS infrastructure. An AWS host manages and controls task allocation, with each trajectory being replayed and evaluated on an independent AWS instance provided by the OSWorld project’s AWS AMI.

\subsection{Agent Framework Ablation}

Currently, our Agent-as-a-Judge is built on MCPMark and demonstrates strong performance compared to LLM-as-a-Judge. To assess the robustness and effectiveness of Agent-as-a-Judge systems, we extend our evaluation to alternative frameworks beyond MCPMark, thereby examining the generalizability of our findings.

As part of our ablation study, we reimplement Agent-as-a-Judge using the ReAct framework, which requires agents to explicitly produce reasoning and actions at each step, in contrast to MCPMark’s more autonomous and implicit reasoning process. Using \textit{gpt-5-mini-low} and \textit{deepseek-v3.2} as the base model, we analyze the impact of different frameworks under the same experimental setting. As shown in Table~\ref{tab:framework_ablation}, although evaluation results vary across frameworks, Agent-as-a-Judge implementations based on both MCPMark and ReAct consistently outperform LLM-as-a-Judge.

\subsection{Agent Model Ablation}
\label{app:agent_model_ablation}

To further examine the generality of the Agent-as-a-Judge setting, we extend the evaluation to four additional judge models spanning both closed-source and open-source families, including Gemini 3 Flash Preview, Claude Sonnet 4.5, Kimi K2.5, and GLM-4.7. Because agentic judging incurs substantially higher evaluation cost, we conduct this analysis on a representative subset of tasks, selecting one subdomain from each of the three domains: Wide (Search), FileSystem (GUI), and PPT (DS). The results are reported in Table~\ref{tab:appendix_agentic_judge_models}. Across all evaluated models, enabling the Agent-as-a-Judge setting consistently improves performance on most subdomains, indicating that agentic judging provides a more effective mechanism for handling complex evaluation scenarios. Among the additional judge models, Gemini 3 Flash Preview achieves the strongest overall performance, while the remaining models also exhibit clear gains in the agentic setting, further supporting the robustness of our conclusion beyond a single judge family.

\subsection{Statistical reliability Analysis}
\label{app:confidence_interval_analysis}

To examine whether the subdomain-level results are sufficiently reliable to support fine-grained conclusions, we report 95\% confidence intervals for all subdomain-level scores based on three independent runs, estimated using the $t$-distribution. Partial results are shown in Table~\ref{tab:subdomain_ci}, and the complete results are provided in the appendix.

The confidence intervals are used to assess two aspects of the subdomain-level findings. First, they verify whether the performance gains brought by Agent-as-a-Judge remain consistent across subdomains despite the highly uneven subset sizes. As shown in Table~\ref{tab:subdomain_ci}, the intervals are shifted upward in most cases for both backbone models, indicating that the improvement is not confined to a small number of isolated subsets. Second, they clarify the extent to which these gains are statistically reliable at a finer granularity. In subdomains with relatively sufficient sample sizes, such as \textsc{Deep}, the intervals are strictly non-overlapping (e.g., GPT-5-mini-low: $[66.18, 70.66]$ vs.\ $[74.89, 76.49]$), which provides clear evidence of statistically reliable improvement. By contrast, in smaller subdomains such as \textsc{PPT} and \textsc{Word}, the intervals are noticeably wider, reflecting higher variance induced by limited sample sizes. Overall, these results show that while the reliability of fine-grained estimates varies across subdomains, the broad upward shift of confidence intervals consistently supports the main conclusion that Agent-as-a-Judge improves subdomain-level performance.

\subsection{Failure Modes Analysis}
\label{app:failure_modes_analysis}

We present a comprehensive analysis of the failure modes encountered by the agent-as-a-judge framework during evaluation. As shown in Table \ref{tab:failure_modes}, we detail the proportional distribution of specific error types across the Search, DS, and GUI domains, and complement our quantitative findings with concrete qualitative examples. Specifically, we categorize the observed failures into four distinct modes: (a) failure to invoke tools or the omission of necessary tool calls; (b) invocation of incorrect tools; (c) misinterpretation of tool outputs; and (d) incorrect reasoning despite the retrieval of accurate evidence.

\begin{table*}[t]
\centering
\resizebox{0.9\textwidth}{!}{
\begin{tabular}{llcccc}
\toprule
\textbf{Domain / Subdomain} & \textbf{Model} & \textbf{(a)} & \textbf{(b)} & \textbf{(c)} & \textbf{(d)} \\
\midrule
\multicolumn{6}{c}{\textit{Search Domain}} \\
\midrule
Deep & deepseek-v3.2 & $\sim$4\% & $\sim$2\% & $\sim$40\% & $\sim$54\% \\
Deep & deepseek-v3.2 w/ thinking & $\sim$1\% & $\sim$1\% & $\sim$30\% & $\sim$68\% \\
Wide & deepseek-v3.2 & $\sim$12\% & negligible & $\sim$57\% & $\sim$31\% \\
Wide & deepseek-v3.2 w/ thinking & $\sim$11\% & negligible & $\sim$55\% & $\sim$34\% \\
\midrule
\multicolumn{6}{c}{\textit{Data Science (DS) Domain}} \\
\midrule
FileSystem & deepseek-v3.2 & $\sim$4.9\% & $\sim$2.4\% & $\sim$56.1\% & $\sim$36.6\% \\
FileSystem & deepseek-v3.2 w/ thinking & negligible & negligible & $\sim$79.2\% & $\sim$20.8\% \\
Postgres & deepseek-v3.2 & $\sim$6.1\% & $\sim$3.0\% & $\sim$48.5\% & $\sim$42.4\% \\
Postgres & deepseek-v3.2 w/ thinking & $\sim$26.7\% & negligible & $\sim$33.3\% & $\sim$40.0\% \\
\midrule
\multicolumn{6}{c}{\textit{GUI Domain}} \\
\midrule
EXCEL & deepseek-v3.2 & $\sim$20.0\% & negligible & $\sim$60.0\% & $\sim$20.0\% \\
EXCEL & deepseek-v3.2 w/ thinking & negligible & negligible & $\sim$80.0\% & $\sim$20.0\% \\
WORD & deepseek-v3.2 & negligible & negligible & $\sim$66.7\% & $\sim$33.3\% \\
WORD & deepseek-v3.2 w/ thinking & negligible & negligible & $\sim$83.3\% & $\sim$16.7\% \\
PPT & deepseek-v3.2 & $\sim$22.2\% & negligible & $\sim$66.7\% & $\sim$11.1\% \\
PPT & deepseek-v3.2 w/ thinking & $\sim$14.3\% & negligible & $\sim$42.9\% & $\sim$42.9\% \\
\bottomrule
\end{tabular}
}
\caption{Proportional distribution of failure modes encountered by the agent-as-a-judge across different domains, subdomains, and models. The error types are defined as follows: \textbf{(a)} failure to invoke tools \& tool omission, \textbf{(b)} invocation of incorrect tools, \textbf{(c)} misinterpretation of tool outputs, and \textbf{(d)} incorrect reasoning despite correct evidence.}
\label{tab:failure_modes}
\end{table*}

\subsection{Additional Metrics}
\label{app:additional_metrics}

As shown in Table~\ref{tab:detailed_metrics}, we have additionally included Precision and Recall, as well as False Positive Rate (FPR) and False Negative Rate (FNR), to provide a comprehensive view of trade-offs in verification.

\begin{table*}[t]
\centering
\resizebox{0.9\textwidth}{!}{%
\begin{tabular}{llcccccccc}
\toprule
\textbf{Model} & \textbf{Agentic} & \textbf{Metric} & \textbf{Wide} & \textbf{Deep} & \textbf{FileSystem} & \textbf{Postgres} & \textbf{PPT} & \textbf{Word} & \textbf{Excel} \\
\midrule
gpt-5-mini-low & $\times$ & FPR & 39.83 & 39.28 & 42.25 & 45.91 & 14.29 & 38.89 & 36.84 \\
gpt-5-mini-low & \checkmark & FPR & 17.65 & 37.00 & 38.03 & 34.59 & 15.88 & 27.78 & 8.77 \\
gpt-5-mini-low & $\times$ & FNR & 36.76 & 37.54 & 34.48 & 26.24 & 66.67 & 55.55 & 35.09 \\
gpt-5-mini-low & \checkmark & FNR & 41.06 & 27.58 & 25.29 & 29.07 & 28.57 & 27.78 & 24.56 \\
gpt-5-mini-low & $\times$ & Precision & 58.65 & 75.65 & 56.19 & 58.89 & 69.80 & 53.37 & 63.89 \\
gpt-5-mini-low & \checkmark & Precision & 74.94 & 79.28 & 61.69 & 65.89 & 81.87 & 72.22 & 89.69 \\
gpt-5-mini-low & $\times$ & Recall & 63.24 & 62.46 & 65.52 & 73.76 & 33.33 & 44.45 & 64.91 \\
gpt-5-mini-low & \checkmark & Recall & 58.94 & 72.42 & 74.71 & 70.92 & 71.43 & 72.22 & 75.44 \\
\midrule
deepseek-v3.2 & $\times$ & FPR & 44.27 & 20.63 & 49.76 & 69.81 & 11.11 & 38.89 & 33.34 \\
deepseek-v3.2 & \checkmark & FPR & 40.56 & 36.05 & 49.29 & 56.60 & 11.08 & 27.78 & 19.30 \\
deepseek-v3.2 & $\times$ & FNR & 30.22 & 49.27 & 30.46 & 11.35 & 53.97 & 25.00 & 28.07 \\
deepseek-v3.2 & \checkmark & FNR & 17.42 & 17.44 & 8.62 & 6.38 & 20.63 & 16.66 & 21.05 \\
deepseek-v3.2 & $\times$ & Precision & 58.51 & 82.77 & 53.27 & 52.98 & 80.51 & 65.45 & 68.62 \\
deepseek-v3.2 & \checkmark & Precision & 64.56 & 81.73 & 60.24 & 59.47 & 87.64 & 75.77 & 80.51 \\
deepseek-v3.2 & $\times$ & Recall & 69.78 & 50.73 & 69.54 & 60.32 & 46.03 & 75.00 & 71.93 \\
deepseek-v3.2 & \checkmark & Recall & 82.58 & 82.56 & 91.37 & 72.60 & 79.37 & 83.34 & 78.95 \\
\bottomrule
\end{tabular}%
}
\caption{Additional evaluation metrics detailing FPR, FNR, Precision, and Recall across different domains.}
\label{tab:detailed_metrics}
\end{table*}

\newpage
\clearpage
\onecolumn

\subsection{Agent-as-a-Judge Failure Cases}
\label{app:failure_cases}

\nolinenumbers
\begin{failure_cases} {Failed to call tools}
**Task**: Find a specific math paper and rename it to \verb|`|answer.html\verb|`|.\\

**What the agent did**: Read 51 paper files but **never renamed any file** — no \verb|`|answer.html\verb|`| was created.\\

**What the judge did**:\\
- Searched for \verb|`|answer.html\verb|`| — confirmed it does **not exist**\\
- The judge explicitly stated: *"The previous agent did not create the required file."*\\
- Only used 4 turns and 608 output tokens before the execution ended\\

**Verdict recorded**: PASS\\

**Category**: **(a) Failed to call tools / system issue** — The judge's own reasoning correctly concluded FAIL, but the recorded verdict was PASS. This suggests either the judge ran out of token budget before emitting a final verdict, or the verdict extraction system parsed an earlier, incorrect output.\\
\end{failure_cases}

\begin{nolinenumbers}
\noindent\begin{minipage}{\textwidth}

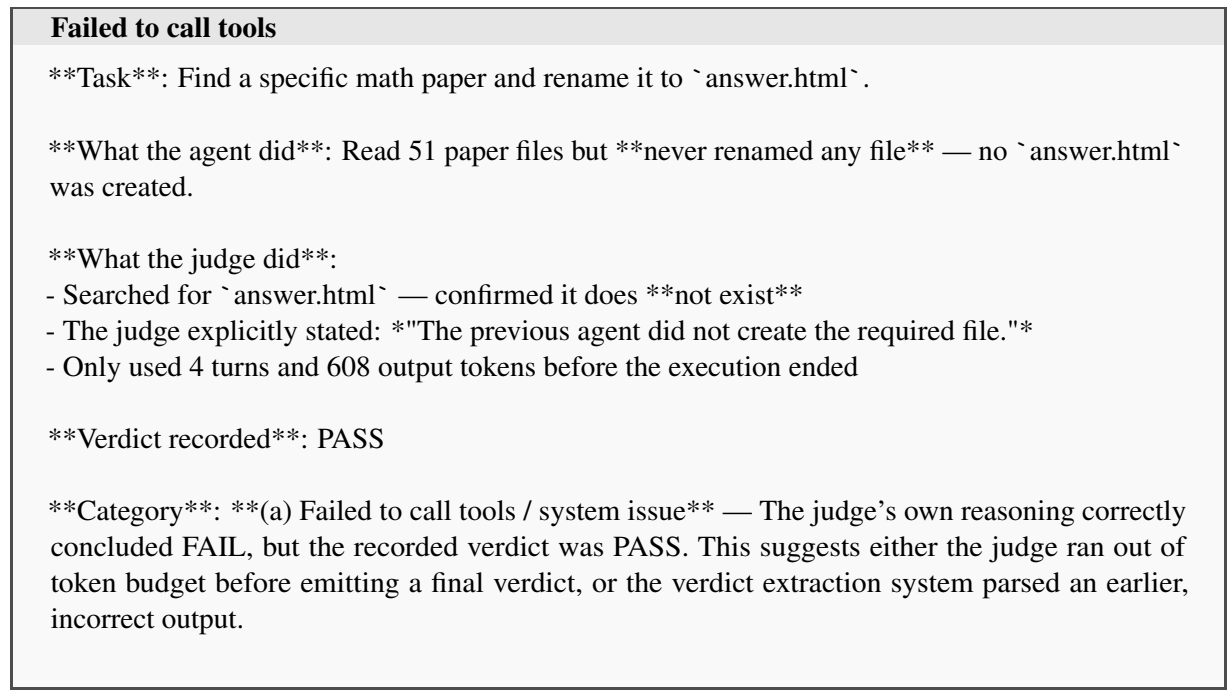
\captionof{figure}{Failed to call tools}
\end{minipage}
\end{nolinenumbers}

\nolinenumbers
\begin{failure_cases} {Misinterpreted tool output}
**Task**: Find and fix the bug in VoteNet's \verb|`|backbone\_module.py\verb|`|.\\

**Ground truth bug**: \verb|`|self.fp2 = PointnetFPModule(mlp=[256,256,256])\verb|`| should be \verb|`|mlp=[512,256,256]\verb|`| (input channels should be 256+256=512).\\

**What the agent did**: Added \verb|`|end\_points['sa3\_inds'] = fps\_inds\verb|`| --- a completely different ``fix'' that doesn't address the actual bug.\\

**What the judge did**:\\
- Read the entire \verb|`|backbone\_module.py\verb|`| (hundreds of lines)\\
- Traced the flow of set abstraction layers (sa1$\rightarrow$sa2$\rightarrow$sa3)\\
- Spent enormous effort analyzing index composition and feature propagation\\
- Never examined whether the \verb|`|fp2\verb|`| layer dimensions were correct\\

**Category**: **(c) Misinterpreted tool output** --- The judge lacked the domain knowledge to identify the actual bug. It was distracted by the agent's red-herring edit and could not independently determine what was wrong.\\
\end{failure_cases}

\begin{nolinenumbers}
\noindent\begin{minipage}{\textwidth}

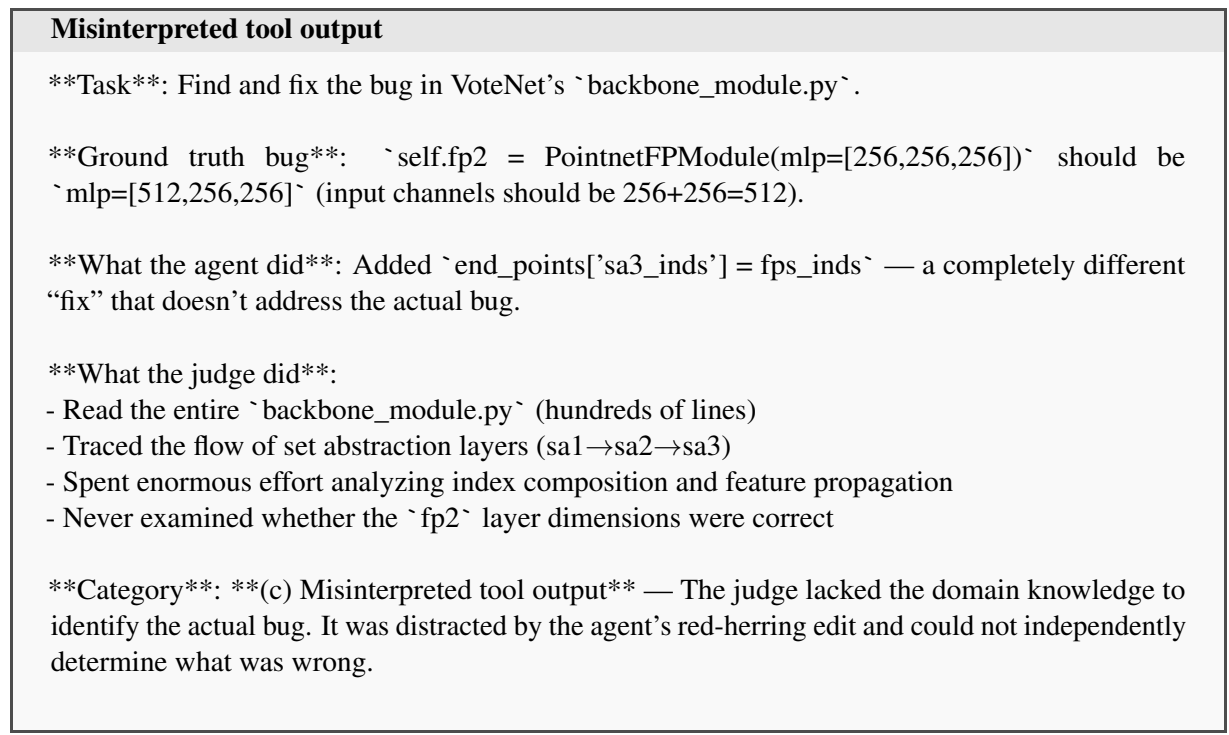
\captionof{figure}{Misinterpreted tool output}
\end{minipage}
\end{nolinenumbers}

\nolinenumbers
\begin{failure_cases} {Correct evidence, wrong reasoning}
**Task**: Find all files with duplicate content and move them to a \verb|`|duplicates/\verb|`| directory.\\

**What the judge did**:\\
- Verified 7 duplicate pairs by reading and hashing all files\\
- Confirmed one copy of each duplicate pair remained in root directory\\
- Discussed both possible interpretations:\\
  - Strict: ALL files with duplicate content go to \verb|`|duplicates/\verb|`| (correct per verify script)\\
  - Lenient: keep one copy in root, move extras to \verb|`|duplicates/\verb|`|\\

**Where it went wrong**:\\

The judge explicitly reasoned through both interpretations, even noting *"if a file has duplicate content, it's not unique"* — which is the verify script's logic. But then concluded:\\

> *"The agent's interpretation is defensible and consistent."*\\

**Category**: **(d) Correct evidence, wrong reasoning** — The judge identified the correct rule but chose not to apply it. The verify script (lines 78-88) explicitly checks that **zero** files from any duplicate group remain in the original directory.\\
\end{failure_cases}

\begin{nolinenumbers}
\noindent\begin{minipage}{\textwidth}
\captionof{figure}{Correct evidence, wrong reasoning}
\end{minipage}
\end{nolinenumbers}

\subsection{Cases Across Domains}
\label{app:domain_cases}
\subsubsection{Search Domain}

\begin{domain_cases}{}
\begin{lstlisting}
Query:

Find 3 movies that premiered in 2024 and had matching theatrical release dates in both mainland China and the United States. For each movie, clearly provide the title, the shared theatrical release date, and credible source links (such as IMDb or official news articles) explicitly confirming these theatrical release dates. Consider only official theatrical release dates—exclude film festival premieres, limited screenings, digital releases, special events, or any other non-standard releases.


Original Trajectory:

# Definitive Findings: Verified Movies with Identical US-China Theatrical Release Dates in 2024

Based on a meticulous review of the available sources, three films from 2024 had matching theatrical release dates in both the United States and mainland China. This section presents a definitive list of these films, complete with their shared release date and the corresponding source verification.

## Movie 1: The Beekeeper

- Shared Theatrical Release Date: January 12, 2024
- Source Verification: The action thriller The Beekeeper opened in the United States on January 12, 2024, as confirmed by a Box Office Mojo entry for the film's domestic release.<sup>1</sup> This date is further corroborated by its Wikipedia page.<sup>2</sup> The film also had its theatrical release in mainland China on the very same day, January 12, 2024.<sup>3</sup>

## Movie 2: Godzilla x Kong: The New Empire

- Shared Theatrical Release Date: March 29, 2024
- Source Verification: This MonsterVerse film had its wide theatrical release in the United States on March 29, 2024, a date confirmed by multiple sources, including its Wikipedia entry <sup>5</sup> and a report from The Numbers.<sup>6</sup> The film also premiered in mainland China on the exact same date. This is explicitly verified by Box Office Mojo's data for the Chinese market <sup>4</sup> and a dedicated page on The Numbers for its China release.<sup>6</sup>

## Movie 3: Deadpool & Wolverine

- Shared Theatrical Release Date: July 26, 2024
- Source Verification: The superhero film Deadpool & Wolverine was theatrically released in the United States on July 26, 2024, as documented in its Wikipedia page.<sup>7</sup> The film's premiere in mainland China also took place on this shared date, a fact confirmed by a YouTube video covering the film's press tour in Shanghai <sup>8</sup> and a fan community post on Reddit that cites news reports about the film's Chinese regulatory approval.<sup>9</sup> The successful day-and-date release of an R-rated film is particularly noteworthy, despite the requirement for "some cuts" to secure its approval in China.<sup>9</sup>

---

## Table 1: 2024 Synchronized US-China Theatrical Releases

| Movie Title | Shared Release Date | U.S. Source | China Source |
| :---- | :---- | :---- | :---- |
| The Beekeeper | January 12, 2024 | 1 | 3 |
| Godzilla x Kong: The New Empire | March 29, 2024 | 5 | 4 |
| Deadpool & Wolverine | July 26, 2024 | 7 | 8 |

### Works cited

1. The Beekeeper - Box Office Mojo, https://www.boxofficemojo.com/release/rl1465155585/  
2. en.wikipedia.org, https://en.wikipedia.org/wiki/The_Beekeeper_(2024_film)  
3. The Beekeeper - Box Office Mojo, https://www.boxofficemojo.com/releasegroup/gr4020130309/  
4. Chinese Box Office For 2024 - Box Office Mojo, https://www.boxofficemojo.com/year/2024/?area=CN  
5. Godzilla x Kong: The New Empire - Wikipedia, https://en.wikipedia.org/wiki/Godzilla_x_Kong:_The_New_Empire  
6. China Box Office for Godzilla x Kong: The New Empire (..., https://www.the-numbers.com/movie/Godzilla-x-Kong-The-New-Empire-(2024)/China  
7. en.wikipedia.org, https://en.wikipedia.org/wiki/Deadpool_%26_Wolverine  
8. Deadpool & Wolverine (2024) China Press Tour 'Fan Event' - YouTube, https://www.youtube.com/watch?v=zqYru1LZ_4Q  
9. Deadpool & Wolverine got the Chinese regulators blessing and and will debut in China's box office on July 26 : r/MarvelStudios_Rumours - Reddit, https://www.reddit.com/r/MarvelStudios_Rumours/comments/1dhnzlr/deadpool_wolverine_got_the_chinese_regulators/


Verification:

I'll evaluate the Single-Row "The Beekeeper" extracted from Rubric-1. Let me follow the verification procedure step by step.

**Step 1 — Identify the Target Rubric Unit**

The Single-Row says "Extracted From Rubric-1". Looking at the Rubric:
1. "Identify films that premiered in 2024 and had consistent theatrical release dates in both mainland China and the United States. Award 1 point for each qualifying film identified, up to a maximum of 3 points."

This is Rubric-1, which requires identifying films that:
- Premiered in 2024
- Had consistent theatrical release dates in both mainland China and the United States

**Step 2 — Trace Source in the Response**

Looking at the Response, "The Beekeeper" appears in:
- "Movie 1: The Beekeeper" section
- Table 1: 2024 Synchronized US-China Theatrical Releases
- Works cited section with references

The Single-Row "The Beekeeper" faithfully reflects the Response content.

**Step 3 — Evidence Verification**

I need to verify the claims about The Beekeeper from the Response references. Let me check the sources provided:

First, let me get a snapshot of the current page to understand the context:
| browser_snapshot {}
I need to navigate to the Response content or check the sources. Let me first navigate to one of the provided URLs to verify the claims about The Beekeeper. I'll start with the Box Office Mojo link for The Beekeeper's domestic release (source 1):
| browser_navigate {"url": "https://www.boxofficemojo.com/release/rl1465155585/"}
Good, this confirms the China release date for The Beekeeper is January 12, 2024. Now let me check the Wikipedia source (source 2) for US release date:
| browser_navigate {"url": "https://en.wikipedia.org/wiki/The_Beekeeper_(2024_film)"}
This confirms the US release date is January 12, 2024. Now let me check the third source for China release verification (source 3):
| browser_navigate {"url": "https://www.boxofficemojo.com/releasegroup/gr4020130309/"}
Perfect! This source clearly shows both Domestic (US) and China release dates as January 12, 2024. Now I need to verify that The Beekeeper premiered in 2024 (which it clearly did based on the January 12, 2024 date).

**Step 4 — Final Verification**

The Single-Row "The Beekeeper" extracted from Rubric-1 makes the claim that this is a film that:
1. Premiered in 2024 (January 12, 2024)
2. Had consistent theatrical release dates in both mainland China and the United States (both January 12, 2024)

The Response provides credible sources that confirm these facts:
- Source 1 (Box Office Mojo): Shows China release date as January 12, 2024
- Source 2 (Wikipedia): Shows US release date as January 12, 2024  
- Source 3 (Box Office Mojo release group): Shows both Domestic (US) and China release dates as January 12, 2024

All claims are factually correct, the Single-Row satisfies ALL requirements of Rubric-1, it faithfully represents its source in the Response, and contains no "N/A".

**Final judgement:** The Single-Row "The Beekeeper" meets all criteria for label [1].

Response: [1]

\end{lstlisting}
\end{domain_cases}

\begin{nolinenumbers}
\noindent\begin{minipage}{\textwidth}
\captionof{figure}{Search Domain Trajectory}
\end{minipage}
\end{nolinenumbers}

\subsubsection{DS Domain}
\begin{domain_cases}{}
\begin{lstlisting}
Query:

Please use FileSystem tools to finish the following task:

Read all the files under current path, extract every time/plan information that clearly indicates 2024, and integrate them into a list and create a file in main directory called `timeline.txt`. Write the timeline in the file in the following format.

### Rules
- If a task only shows month without day, use the 1st day of that month
- If a task only shows year without month and day, skip it.
- If a file shows multiple tasks on the same date, count only once per date

### Output Format
- Each line format: `file_path:time`
    - `file_path`: The file path where this time information appears (**relative to the current path**)
    - `time`: Specific time, if it's a time period, write the start time (YYYY-MM-DD)

### Sorting Requirements
- Sort by chronological order


Original Trajectory:

| list_allowed_directories {}
| directory_tree {"path": "/home/<user>/.mcpmark_backups/backup_<task_name>_<id>"}
| read_multiple_files {"paths": ["/home/<user>/.mcpmark_backups/backup_<task_name>_<id>/exp_logs/aug/augmentation_log.txt","..."]}
| write_file {"path": "/home/<user>/.mcpmark_backups/backup_<task_name>_<id>/timeline.txt","content": "exp_logs/..."}
Task completed


Verification:

## Evaluation Steps

### Step 1
- **Action Type:** `list_allowed_directories`
- **Reasoning (pre-action content):**  
  I'll verify whether the task was completed successfully by checking the current state and examining the created timeline file and its contents. First, let me see what directories are accessible:
- **Action Call:**  
  `| list_allowed_directories {}`
- **Observation:**  
  Accessible directories were listed.

---

### Step 2
- **Action Type:** `directory_tree`
- **Target Path (Anonymized):** `/path/to/backup_root`
- **Reasoning (pre-action content):**  
  Now let me examine the directory structure to see what files exist:
- **Action Call:**  
  `| directory_tree {"path": "/path/to/backup_root"}`
- **Observation:**  
  The directory tree shows that `timeline.txt` exists in the main directory.

---

### Step 3
- **Action Type:** `read_text_file`
- **Target Path (Anonymized):** `/path/to/backup_root/timeline.txt`
- **Reasoning (pre-action content):**  
  Great! I can see that timeline.txt exists in the main directory. Now let me check its contents to verify if it was created correctly according to the task requirements:
- **Action Call:**  
  `| read_text_file {"path": "/path/to/backup_root/timeline.txt"}`
- **Observation:**  
  `timeline.txt` contents were retrieved for verification.

---

### Step 4
- **Action Type:** `read_multiple_files`
- **Target Files (Anonymized):**
  - `/path/to/backup_root/exp_logs/project_2/analysis_report.md`
  - `/path/to/backup_root/learning/2024/learning_progress.csv`
  - `/path/to/backup_root/exp_logs/experiment_summary.md`
- **Reasoning (pre-action content):**  
  Now I need to verify that the timeline.txt file meets all the requirements. Let me check a few key files to see if the extracted dates are correct. First, let me examine some of the files mentioned in the timeline to see if they actually contain 2024 date information:
- **Action Call:**  
  `| read_multiple_files {"paths": ["/path/to/backup_root/exp_logs/project_2/analysis_report.md","/path/to/backup_root/learning/2024/learning_progress.csv","/path/to/backup_root/exp_logs/experiment_summary.md"]}`
- **Observation:**  
  The files do contain 2024 date information.

---

### Step 5
- **Action Type:** `Manual Verification (Chronological Sorting)`
- **Reasoning (pre-action content):**  
  Good! I can see that the files do contain 2024 date information. Now let me check if the timeline is sorted chronologically. Looking at the timeline.txt content, I can see it starts with 2024-01-01 entries and progresses through the year. Let me verify the sorting more carefully by checking a few entries:
  1. First entry: exp_logs/project_2/analysis_report.md:2024-01-01
  2. Second entry: learning/2024/learning_progress.csv:2024-01-01 (same date, different file)
  3. Third entry: exp_logs/experiment_summary.md:2024-01-05 (later date)
  4. Last entry: learning/2024/learning_progress.csv:2024-12-01 (December)
  The sorting appears to be chronological.
- **Observation:**  
  Chronological sorting appears correct.

---

...

---

### Step 21
- **Action Type:** `read_multiple_files`
- **Target Files (Anonymized):**
  - `/path/to/backup_root/exp_logs/project_1/README.md`
  - `/path/to/backup_root/exp_logs/project_1/data.csv`
  - `/path/to/backup_root/exp_logs/project_1/model.py`
  - `/path/to/backup_root/exp_logs/project_2/data_analysis.py`
- **Reasoning (pre-action content):**  
  However, I should check if there are any other date formats that might have been missed. Let me quickly scan through all text files for any date patterns:
- **Action Call:**  
  `| read_multiple_files {"paths": ["/path/to/backup_root/exp_logs/project_1/README.md","/path/to/backup_root/exp_logs/project_1/data.csv","/path/to/backup_root/exp_logs/project_1/model.py","/path/to/backup_root/exp_logs/project_2/data_analysis.py"]}`
- **Observation:**  
  These files were checked for missed 2024 date patterns.

---

### Step 22
- **Action Type:** `Manual Verification (Final Coverage Check)`
- **Reasoning (pre-action content):**  
  These files don't contain any 2024 dates, so they shouldn't be in the timeline.
- **Observation:**  
  No additional missing 2024 entries were identified from this final scan.

---

## Final judgement

- **judgement:** `true`
- **Verdict:** `PASS`
- **Confidence:** `High`

### Detailed Reasoning (Original Information Preserved)

Based on my thorough verification, I can confirm that:

1. The timeline.txt file was created in the main directory  
2. All 2024 date information from all files was extracted  
3. Month-only dates were converted to the 1st day of the month  
4. Year-only entries were skipped  
5. No duplicate dates for the same file appear  
6. The format is correct: file_path:time with relative paths  
7. Sorting is chronological  
8. Time periods use the start time  

The task appears to have been completed successfully according to all requirements.

**VERDICT: PASS**
\end{lstlisting}
\end{domain_cases}

\begin{nolinenumbers}
\noindent\begin{minipage}{\textwidth}
\captionof{figure}{DS Domain Trajectory}
\end{minipage}
\end{nolinenumbers}

\subsubsection{GUI Domain}
\begin{domain_cases}
\begin{lstlisting}
Query:

Move the image to the right side on Slide 2.

Original Trajectory:

{"step_num": 1, "action_timestamp": "20250722@212748", "action": {"name": "computer", "input": {"action": "left_click", "coordinate": [116, 263]}, "id": "toolu_bdrk_015e8fhjpZH3AJ58Tn3znPT9", "action_type": "tool_use", "command": "pyautogui.click(174, 394)\n"}, "reward": 0, "done": false, "info": {}, "screenshot_file": "step_1_20250722@212748.png"}
{"step_num": 2, "action_timestamp": "20250722@212819", "action": {"name": "computer", "input": {"action": "left_click", "coordinate": [325, 364]}, "id": "toolu_bdrk_01H4ycd7eDAZ3jHP7JvHmbMu", "action_type": "tool_use", "command": "pyautogui.click(487, 546)\n"}, "reward": 0, "done": false, "info": {}, "screenshot_file": "step_2_20250722@212819.png"}
{"step_num": 3, "action_timestamp": "20250722@212830", "action": {"name": "computer", "input": {"action": "left_click_drag", "start_coordinate": [325, 364], "coordinate": [700, 364]}, "id": "toolu_bdrk_01F4DrWba3tW1K3cHhh9uSSx", "action_type": "tool_use", "command": "pyautogui.dragTo(1050, 546, duration=0.5)\n"}, "reward": 0, "done": false, "info": {}, "screenshot_file": "step_3_20250722@212830.png"}
{"step_num": 4, "action_timestamp": "20250722@212934", "action": {"name": "computer", "input": {"action": "left_click", "coordinate": [400, 500]}, "id": "toolu_bdrk_01A32q84UHBnfFtevEpDTCfp", "action_type": "tool_use", "command": "pyautogui.click(600, 750)\n"}, "reward": 0, "done": false, "info": {}, "screenshot_file": "step_4_20250722@212934.png"}
{"step_num": 5, "action_timestamp": "20250722@212952", "action": "DONE", "reward": 0, "done": true, "info": {"done": true}, "screenshot_file": "step_5_20250722@212952.png"}


Verification:

# Evaluation Steps and Final judgement

## Evaluation Steps

### Step 1
- **Action Type:** `VIEW_TRAJECTORY_STEP`
- **Viewed Step:** `5`
- **Action String:** `{"action_type": "VIEW_TRAJECTORY_STEP", "step": 5}`
- **Reasoning:**  
  I need to first review the original trajectory (mandatory) and then gather evidence. Starting with `VIEW_TRAJECTORY_STEP` for the final step to inspect the environment state after trajectory execution. This provides an initial view of Slide 2 and the image position.
- **Observation:**  
  `Trajectory Step 5:`  
  `Action: DONE`  
  `Accessibility Tree: <Step5 Accessibility Tree>`

---

### Step 2
- **Action Type:** `CLICK`
- **Coordinates:** `(930, 412)`
- **Action String:** `{"action_type": "CLICK", "x": 930, "y": 412}`
- **Reasoning:**  
  I need to interact with the live environment to gather evidence about the current state. The accessibility tree from Step 5 shows an image (`"Picture 3"`) at position `(930, 412)` with size `(256, 256)`. I click the image to select it and verify whether it is on Slide 2 and positioned on the right side as required. This may also reveal selection handles or properties.
- **Observation:**  
  `<Accessibility Tree>`

---

### Step 3
- **Action Type:** `VIEW_TRAJECTORY_STEP`
- **Viewed Step:** `1`
- **Action String:** `{"action_type": "VIEW_TRAJECTORY_STEP", "step": 1}`
- **Reasoning:**  
  I need more evidence about the trajectory execution. The current accessibility tree shows the image at `(930, 412)`, which appears to be on the right side of the slide (slide width is `1123`, image left edge at `930`, right edge at `1186`). However, I also need to verify that this is Slide 2 (the status bar says `"Slide 2 of 2"`) and determine whether the image was moved from a different position. I examine Step 1 to see the initial state before any actions.
- **Observation:**  
  `Trajectory Step 1:`  
  `Action: {'name': 'computer', 'input': {'action': 'left_click', 'coordinate': [116, 263]}, 'id': 'toolu_bdrk_015e8fhjpZH3AJ58Tn3znPT9', 'action_type': 'tool_use', 'command': 'pyautogui.click(174, 394)\n'}`  
  `Accessibility Tree: <Step1 Accessibility Tree>`

---

### Step 4
- **Action Type:** `VIEW_TRAJECTORY_STEP`
- **Viewed Step:** `3`
- **Action String:** `{"action_type": "VIEW_TRAJECTORY_STEP", "step": 3}`
- **Reasoning:**  
  I need to inspect Step 3, where the drag action occurred. Step 1 shows the image initially at `(364, 415)` on the left side of the slide. Step 5 shows the image at `(930, 412)` on the right side. Step 3 contains a drag action from `(325, 364)` to `(700, 364)`. I inspect the state after Step 3 to confirm that the drag moved the image.
- **Observation:**  
  `Trajectory Step 3:`  
  `Action: {'name': 'computer', 'input': {'action': 'left_click_drag', 'start_coordinate': [325, 364], 'coordinate': [700, 364]}, 'id': 'toolu_bdrk_01F4DrWba3tW1K3cHhh9uSSx', 'action_type': 'tool_use', 'command': 'pyautogui.dragTo(1050, 546, duration=0.5)\n'}`  
  `Accessibility Tree: <Step3 Accessibility Tree>`

---

### Step 5
- **Action Type:** `CLICK`
- **Coordinates:** `(73, 204)`
- **Action String:** `{"action_type": "CLICK", "x": 73, "y": 204}`
- **Reasoning:**  
  I interact with the live environment again to verify the current state. The accessibility tree from Step 3 shows the image at `(930, 412)`, matching the final position. I click the `"Slides View"` document-frame at `(73, 204)` to potentially view slide thumbnails and confirm that the current slide is Slide 2.
- **Observation:**  
  `<Accessibility Tree>`

---

### Step 6
- **Action Type:** `CLICK`
- **Coordinates:** `(930, 412)`
- **Action String:** `{"action_type": "CLICK", "x": 930, "y": 412}`
- **Reasoning:**  
  I select the image again to verify that it remains on the right side. The current accessibility tree shows the image at `(930, 412)`, and the status bar shows `"Slide 2 of 2"`. Clicking the image also helps confirm its selection state and position relative to the slide.
- **Observation:**  
  `<Accessibility Tree>`

---

## Final judgement

- **judgement:** `true`
- **Confidence:** `High`

### Detailed Reasoning

The task instruction is: **“Move the image to the right side on Slide 2.”**  
Based on the gathered evidence from both trajectory inspection and live interactions, the task was successfully completed.

1. **Initial Position (Step 1)**  
   The `VIEW_TRAJECTORY_STEP` for Step 1 shows image `"Picture 3"` at position `(364, 415)`, which is on the **left side** of the slide.  
   The slide presentation area spans from `x = 273` to `x = 1396` (width `1123`), so `x = 364` is clearly left-of-center.

2. **Final Position (Live Interaction)**  
   In the live environment, after clicking the image, the accessibility tree shows the image at `(930, 412)` with size `(256, 256)`.  
   This places:
   - **Left edge:** `x = 930`
   - **Right edge:** `x = 1186`  
   This is clearly on the **right side** of the slide.

3. **Slide Verification**  
   The status bar displays **“Slide 2 of 2”** in both the live environment and trajectory steps, confirming the image is on **Slide 2**, as required.

4. **Trajectory Execution Evidence**  
   The trajectory includes a drag action in Step 3 from `(325, 364)` to `(700, 364)`, consistent with moving the image horizontally to the right.  
   The corresponding command:
   - `pyautogui.dragTo(1050, 546, duration=0.5)`  
   also indicates a rightward movement.  
   Additionally, Step 3’s accessibility tree already shows the image at `(930, 412)`, confirming the drag successfully repositioned it.

5. **Selection Confirmation**  
   After clicking the image in the live environment, the status bar updates to indicate that image `"Picture 3"` is selected, and the properties panel shows the **Image** toggle/button active. This confirms the UI is recognizing the correct object.

6. **Right-Side Assessment**  
   The slide center is approximately at `x = 834` (`273 + 1123/2`). The image’s left edge at `x = 930` is to the **right of center**, and the image occupies the rightmost portion of the slide, satisfying the requirement of being “on the right side.”

### Conclusion
The image was successfully moved from the left side to the right side **on Slide 2**, and the trajectory fulfills the task requirement.
\end{lstlisting}
\end{domain_cases}

\begin{nolinenumbers}
\noindent\begin{minipage}{\textwidth}
\captionof{figure}{GUI Domain Trajectory}
\end{minipage}
\end{nolinenumbers}

\subsection{Prompts}
\label{app:evaluation_prompts}
\subsubsection{Search Domain}

\nolinenumbers
\begin{llm_judge_search} {LLM-as-a-Judge Prompt (Wide)}
\# LLM-as-a-Judge Task\\

You will verify the factual correctness of ONE Single-Row extracted from a Response.\\
You MUST follow all instructions precisely and avoid any interpretation that goes beyond what is explicitly required.\\

---------------------------------------------------\\
\#\# Provided Inputs\\

\#\#\# Query\\

[The Start of Query]\\

\{query\}\\

[The End of Query]\\

\#\#\# Response\\

[The Start of Response]\\

\{response\}\\

[The End of Response]\\

\#\#\# Single-Row (One extracted item)\\

[The Start of Single-Row]\\

\{single\_row\}\\

[The End of Single-Row]\\

---------------------------------------------------\\
\#\# Core Principle\\

Your task is NOT to evaluate whether the Single-Row answers the entire Query. Your task IS ONLY to evaluate whether THIS Single-Row:\\

1. Correctly reflects the part of the Response it was extracted from;\\
2. Is factually correct according to your knowledge.\\

You MUST ignore completeness. You MUST evaluate only the correctness of THIS Single-Row's own claims.\\

---------------------------------------------------\\
\#\# Role and High-Level Objective\\

Act as a neutral, impartial judge.\\

---------------------------------------------------\\
\#\# Step-by-Step Evaluation Procedure\\

\#\#\# Step 1 — Initial Read\\
- Read the Query, Response, and Single-Row fully.\\

\#\#\# Step 2 — Trace Source in the Response\\
- Locate the corresponding content in the Response.\\
- Confirm the Single-Row faithfully reflects the Response.\\

\#\#\# Step 3 — Verify Each Claim Individually\\
- Focus strictly on content relevant to the Query.\\
- Do not let style, language, or level of detail influence judgement.\\
- Treat "N/A", "NA", or "Data Unavailable" as problematic unless explicitly allowed.\\
- Verify every field if the item contains multiple attributes (e.g., in a table row).\\
- Numeric values must strictly match authoritative sources—no tolerance allowed.\\

\#\#\# Step 4 — Record Verification Results\\
For every claim in the Single-Row record:\\
- Supported / Contradicted / Not verifiable\\

---------------------------------------------------\\
\#\# Final judgement Rules\\

A Single-Row is labeled **[1]** ONLY IF:\\
1. It contains at least one claim relevant to the Query; AND\\
2. Every claim is factually correct; AND\\
3. It faithfully represents the Response.\\

Otherwise, assign **[0]**.\\

\#\#\# If ANY claim is:\\
- incorrect, OR\\
- missing, OR\\
- unverifiable\\
→ assign **[0]**\\

---------------------------------------------------\\
\#\# Strict Requirements\\

- You MUST NOT add or infer information beyond what is provided.\\

---------------------------------------------------\\
\#\# Output Format (Strict)\\

Your final output must be exactly one line:\\

Response: [1]\\
or\\
Response: [0]\\

Nothing else.
\end{llm_judge_search}

\begin{nolinenumbers}
\noindent\begin{minipage}{\textwidth}
\captionof{figure}{Search Domain (Wide) LLM-as-a-Judge Prompt}
\end{minipage}
\end{nolinenumbers}

\begin{llm_judge_search} {LLM-as-a-Judge Prompt (Deep)}
\# LLM-as-a-Judge Task\\

You will verify the factual correctness and rubric compliance of ONE Single-Row extracted from a Response.\\
You MUST follow all instructions precisely and avoid any interpretation that goes beyond what is explicitly required.\\

---------------------------------------------------\\
\#\# Provided Inputs\\

\#\#\# Query\\

[The Start of Query]\\

\{query\}\\

[The End of Query]\\

\#\#\# Response\\

[The Start of Response]\\

\{response\}\\

[The End of Response]\\

\#\#\# Rubric\\

[The Start of Rubric]\\

\{rubric\}\\

[The End of Rubric]\\

\#\#\# Single-Row (One extracted item)\\

[The Start of Single-Row]\\

\{single\_row\}\\

[The End of Single-Row]\\

---------------------------------------------------\\
\#\# Core Principle\\

**Your task is NOT to evaluate whether the Single-Row answers the entire Query. Your task IS ONLY to evaluate whether THIS extracted Single-Row:**\\

1. **Directly corresponds to exactly ONE scoring unit in the Rubric;**\\
2. **Correctly reflects the part of the Response it was extracted from;**\\
3. **Is factually correct according to your knowledge;**\\
4. **Meets ALL requirements of the corresponding Rubric scoring unit;**\\
5. **Contains at least one claim relevant to the Query.**\\

You MUST ignore completeness.\\
You MUST evaluate only the correctness of THIS Single-Row's own claims.\\

---------------------------------------------------\\
\#\# Structural Constraints (Very Important)\\

1. **Single-Row = One scoring unit**\\
    - The "Extracted From Rubric" ID in the Single-Row must match exactly one scoring unit in the overall Rubric.\\

2. **No mixing**\\
    - The Single-Row must not combine multiple rubric units into one claim.\\

3. **Strict multi-condition requirement**\\
    - If the matched Rubric scoring unit contains multiple requirements, **the Single-Row must satisfy ALL of them. Partial fulfillment = label [0].**\\

4. **"N/A" rule (mandatory)**\\
    - If the Single-Row contains "N/A", **you must immediately label it as [0], regardless of any other information.**\\

---------------------------------------------------\\
\#\# Distinguishing Requirement Types (CRITICAL)\\

You MUST determine whether the matched Rubric scoring unit requires:\\

\#\#\# A. Factual content\\
(e.g., a person's name, a date, a fact, an explanation)\\
→ You MAY use your own prior knowledge to verify factual claims.\\

\#\#\# B. A URL, page link, or reference link\\
→ You MUST evaluate the correctness of a link using ALL of the following sources:\\
- your own prior knowledge,\\
- your ability to judge whether a link is plausible, internally consistent, or matches well-known link patterns.\\

→ You MUST NOT:\\
- generate a new link,\\
- search for the correct link,\\
- fix the link,\\
- replace anything missing,\\
- assume a link is correct based solely on its appearance.\\

You MUST NOT judge a link as correct ONLY based on formatting, domain familiarity, or similarity to known sites. You MAY use your internal knowledge to assess whether the provided link is likely to be correct.\\

If the Rubric requires a link and the Single-Row does not provide one → label [0].\\

---------------------------------------------------\\
\#\# Verification Procedure\\

\#\#\# Step 1 — Identify the Target Rubric Unit\\
- Locate the scoring unit whose index matches "Extracted From Rubric" in the Single-Row.\\
- All evaluation must be restricted to the requirements of that scoring unit.\\

\#\#\# Step 2 — Trace Source in the Response\\
- Identify where in the Response the Single-Row was extracted from.\\
- Confirm the Single-Row content faithfully reflects the Response.\\

\#\#\# Step 3 — Evidence Verification\\
You MUST:\\
- Follow all explicit references in the Response that are relevant to the Single-Row, including URLs, footnotes (e.g., [\textasciicircum 1], <sup>1</sup>), or entries in "Work Cited" / "Key Citations".\\
- Validate every claim in the Single-Row that attempts to satisfy the Rubric.\\

\#\#\# Step 4 — Handling "Unverifiable"\\
If a claim remains unverifiable after:\\
- reviewing all Response references, and\\
- applying your own prior knowledge and reasoning,\\

then mark the claim as Not verifiable and assign label [0].\\

A link is Not verifiable ONLY if:\\
- you have no relevant knowledge of its expected structure, AND\\
- the Response provides no context that helps determine correctness.\\

---------------------------------------------------\\
\#\# Final judgement Rules\\

Label the Single-Row as **[1]** only if:\\

1. It makes at least one claim relevant to the Query AND\\
2. Every claim is factually correct AND\\
3. It satisfies ALL requirements of the matched Rubric scoring unit AND\\
4. It faithfully represents its source in the Response AND\\
5. No part of the claim is "N/A".\\

Otherwise, assign **[0]**.\\

---------------------------------------------------\\
\#\# Output Format (Strict)\\

You MUST output only:\\

- Response: [1]\\

or\\

- Response: [0]\\

Nothing else.
\end{llm_judge_search}

\begin{nolinenumbers}
\noindent\begin{minipage}{\textwidth}
\captionof{figure}{Search Domain (Deep) LLM-as-a-Judge Prompt}
\end{minipage}
\end{nolinenumbers}

\begin{agent_judge_search} {Agent-as-a-Judge Prompt (Wide)}
\# Agent-as-a-Judge Task\\

You will verify the factual correctness of ONE Single-Row extracted from a Response.\\
You MUST follow all instructions precisely and avoid any interpretation that goes beyond what is explicitly required.\\

---------------------------------------------------\\
\#\# Provided Inputs\\

\#\#\# Query\\

[The Start of Query]\\

\{query\}\\

[The End of Query]\\

\#\#\# Response\\

[The Start of Response]\\

\{response\}\\

[The End of Response]\\

\#\#\# Single-Row (One extracted item)\\

[The Start of Single-Row]\\

\{single\_row\}\\

[The End of Single-Row]\\

---------------------------------------------------\\
\#\# Core Principle\\

Your task is NOT to evaluate whether the Single-Row answers the entire Query. Your task IS ONLY to evaluate whether THIS Single-Row:\\

1. Correctly reflects the part of the Response it was extracted from;\\
2. Is factually correct according to verification tools.\\

You MUST ignore completeness. You MUST evaluate only the correctness of THIS Single-Row's own claims.\\

---------------------------------------------------\\
\#\# Role and High-Level Objective\\

Act as a neutral, impartial judge. All factual checks must be performed using Playwright MCP tools.\\

---------------------------------------------------\\
\#\# Step-by-Step Evaluation Procedure\\

\#\#\# Step 1 — Initial Read\\
- Read the Query, Response, and Single-Row fully.\\

\#\#\# Step 2 — Trace Source in the Response\\
- Locate the corresponding content in the Response.\\
- Confirm the Single-Row faithfully reflects the Response.\\

\#\#\# Step 3 — Verify Each Claim Individually with Playwright MCP Tools\\
- Focus strictly on content relevant to the Query.\\
- Do not let style, language, or level of detail influence judgement.\\
- Treat "N/A", "NA", or "Data Unavailable" as problematic unless explicitly allowed.\\
- Follow explicit references in the Response: URLs, citations, or footnotes (e.g., [\textasciicircum 1], <sup>1</sup>).\\
- Open referenced pages with Playwright MCP tools and locate supporting evidence.\\
- Primarily rely on references in the Response; if none exist, search authoritative sources using the tools.\\
- If a reference is unreachable, paywalled, or does not support the claim, mark accordingly.\\
- Verify every field if the item contains multiple attributes (e.g., in a table row).\\
- Numeric values must strictly match authoritative sources—no tolerance allowed.\\

\#\#\# Step 4 — Record Verification Results\\
For every claim in the Single-Row record:\\
- Supported / Contradicted / Not verifiable\\

---------------------------------------------------\\
\#\# Final judgement Rules\\

A Single-Row is labeled **[1]** ONLY IF:\\
1. It contains at least one claim relevant to the Query; AND\\
2. Every claim is factually correct; AND\\
3. It faithfully represents the Response.\\

Otherwise, assign **[0]**.\\

\#\#\# If ANY claim is:\\
- incorrect, OR\\
- missing, OR\\
- unverifiable\\
→ assign **[0]**\\

---------------------------------------------------\\
\#\# Strict Requirements\\

- You MUST use Playwright MCP tools for all factual verification.\\
- You MUST NOT add or infer information beyond what is provided.\\

---------------------------------------------------\\
\#\# Output Format (Strict)\\

Your final output must be exactly one line:\\

Response: [1]\\
or\\
Response: [0]\\

Nothing else.\\
\end{agent_judge_search}

\begin{nolinenumbers}
\noindent\begin{minipage}{\textwidth}
\captionof{figure}{Search Domain (Wide) Agent-as-a-Judge Prompt}
\end{minipage}
\end{nolinenumbers}

\begin{agent_judge_search} {Agent-as-a-Judge Prompt (Deep)}
\# Agent-as-a-Judge Task\\

You will verify the factual correctness and rubric compliance of ONE Single-Row extracted from a Response.\\
You MUST follow all instructions precisely and avoid any interpretation that goes beyond what is explicitly required.\\

---------------------------------------------------\\
\#\# Provided Inputs\\

\#\#\# Query\\

[The Start of Query]\\

\{query\}\\

[The End of Query]\\

\#\#\# Response\\

[The Start of Response]\\

\{response\}\\

[The End of Response]\\

\#\#\# Rubric\\

[The Start of Rubric]\\

\{rubric\}\\

[The End of Rubric]\\

\#\#\# Single-Row (One extracted item)\\

[The Start of Single-Row]\\

\{single\_row\}\\

[The End of Single-Row]\\

---------------------------------------------------\\
\#\# Core Principle\\

**Your task is NOT to evaluate whether the Single-Row answers the entire Query. Your task IS ONLY to evaluate whether THIS extracted Single-Row:**\\

1. **Directly corresponds to exactly ONE scoring unit in the Rubric;**\\
2. **Correctly reflects the part of the Response it was extracted from;**\\
3. **Is factually correct according to verification tools;**\\
4. **Meets ALL requirements of the corresponding Rubric scoring unit;**\\
5. **Contains at least one claim relevant to the Query.**\\

You MUST ignore completeness.\\
You MUST evaluate only the correctness of THIS Single-Row's own claims.\\

---------------------------------------------------\\
\#\# Structural Constraints (Very Important)\\

1. **Single-Row = One scoring unit**\\
    - The "Extracted From Rubric" ID in the Single-Row must match exactly one scoring unit in the overall Rubric.\\

2. **No mixing**\\
    - The Single-Row must not combine multiple rubric units into one claim.\\

3. **Strict multi-condition requirement**\\
    - If the matched Rubric scoring unit contains multiple requirements, **the Single-Row must satisfy ALL of them. Partial fulfillment = label [0].**\\

4. **"N/A" rule (mandatory)**\\
    - If the Single-Row contains "N/A", **you must immediately label it as [0], regardless of any other information.**\\

---------------------------------------------------\\
\#\# Distinguishing Requirement Types (CRITICAL)\\

You MUST determine whether the matched Rubric scoring unit requires:\\

\#\#\# A. Factual content\\
(e.g., a person's name, a date, a fact, an explanation)\\
→ You MAY search authoritative sources using Playwright MCP tools.\\

\#\#\# B. A URL, page link, or reference link\\
→ You MUST evaluate ONLY the correctness of the link provided.\\
→ You MUST NOT:\\
- generate a new link,\\
- search for the correct link,\\
- fix the link,\\
- replace anything missing.\\

If the Rubric requires a link and the Single-Row does not provide one → label [0].\\

---------------------------------------------------\\
\#\# Verification Procedure\\

\#\#\# Step 1 — Identify the Target Rubric Unit\\
- Locate the scoring unit whose index matches "Extracted From Rubric" in the Single-Row.\\
- All evaluation must be restricted to the requirements of that scoring unit.\\

\#\#\# Step 2 — Trace Source in the Response\\
- Identify where in the Response the Single-Row was extracted from.\\
- Confirm the Single-Row content faithfully reflects the Response.\\

\#\#\# Step 3 — Evidence Verification (Using Tools)\\
You MUST:\\

- Follow all explicit references in the Response that are relevant to the Single-Row, including URLs, footnotes (e.g., [\textasciicircum 1], <sup>1</sup>), or entries in "Work Cited" / "Key Citations".\\
- Open referenced pages using Playwright MCP tools.\\
- Validate every claim in the Single-Row that attempts to satisfy the Rubric.\\

If Response references are insufficient:\\
- Only if the Rubric requires *factual content*, you may perform searches.\\
- If the Rubric requires *a link*, you MUST NOT search for any missing link.\\

\#\#\# Step 4 — Handling "Unverifiable"\\
If a claim remains unverifiable after:\\
- reviewing all Response references, and\\\\
- performing allowed searches,\\

then mark the claim as **Not verifiable** and assign label [0].\\

---------------------------------------------------\\
\#\# Final judgement Rules\\

Label the Single-Row as **[1]** only if:\\

1. It makes at least one claim relevant to the Query AND\\
2. Every claim is factually correct AND\\
3. It satisfies ALL requirements of the matched Rubric scoring unit AND\\
4. It faithfully represents its source in the Response AND\\
5. No part of the claim is "N/A".\\

Otherwise, assign **[0]**.\\

---------------------------------------------------\\
\#\# Output Format (Strict)\\

You MUST output only:\\

- Response: [1]\\

or\\

- Response: [0]\\

Nothing else.
\end{agent_judge_search}

\begin{nolinenumbers}
\noindent\begin{minipage}{\textwidth}
\captionof{figure}{Search Domain (Deep) Agent-as-a-Judge Prompt}
\end{minipage}
\end{nolinenumbers}

\linenumbers

\subsubsection{DS Domain}

\nolinenumbers
\begin{llm_judge_ds} {LLM-as-a-Judge Prompt}
    You are an impartial evaluator. Your job is to determine whether the given Agent Trajectory successfully completes the Task Description. \\

You must follow these rules strictly:\\

(1) Base your judgement only on the information provided in the Task Description and the Agent Trajectory.\\
(2) Evaluate whether every requirement in the Task Description is fully satisfied.\\
(3) Your answer must be binary: `PASS' or `bangbangFAIL'.\\
(4) Before giving the binary result, you may provide chain-of-thought or step-by-step internal reasoning.\\
(5) Do not propose improvements, fixes, or technical solutions—simply judge.\\

Input Format:\\

[TASK DESCRIPTION]\\

\{task\}\\

[AGENT TRAJECTORY]\\

\{trajectory\} \\

Judging Criteria:
Mark the trajectory as PASS only if all required steps are present, all computations or file operations are correct, the output format exactly matches the specification, no required content is missing, and no extra or disallowed content is added. If any requirement is unmet, the result must be FAIL.
\\

Output Format (Strict):\\

RESULT: PASS or FAIL
\end{llm_judge_ds}

\begin{nolinenumbers}
\noindent\begin{minipage}{\textwidth}
\captionof{figure}{DS Domain LLM-as-a-Judge Prompt}
\end{minipage}
\end{nolinenumbers}

\begin{agent_judge_ds} {Agent-as-a-Judge Prompt}
\# Task Description \\

\{task\_description\} \\

\# Previous Agent Trajectory (Summary)\\

The following is a summary of what the previous agent did:\\

\{trajectory\_text\}\\

\# Your Verification Task
The previous agent claimed to have completed the task above. Your job is to verify whether the task was actually completed successfully. \\

Use the available tools to:\\
1. Check the current state of the environment\\
2. Verify that the expected outputs/changes exist\\
3. Confirm the correctness of any created or modified resources\\

After your investigation, provide your final verdict:\\
- "VERDICT: PASS" if the task was completed successfully\\
- "VERDICT: FAIL" if the task was not completed or was completed incorrectly\\

Be thorough and check all requirements mentioned in the task description.
\end{agent_judge_ds}

\begin{nolinenumbers}
\noindent\begin{minipage}{\textwidth}
\captionof{figure}{DS Domain Agent-as-a-Judge Prompt}
\end{minipage}
\end{nolinenumbers}

\linenumbers

\subsubsection{GUI Domain}

The relatively long prompts in the GUI domain do not mainly reflect task-specific instruction complexity, but rather implementation constraints. Specifically, because OSWorld does not provide an official MCP server, we introduced a lightweight custom adapter to integrate it into our framework. As a result, tool definitions, parameter constraints, and return schemas—components that would normally be enforced by a structured backend interface—must be explicitly specified in the prompt to ensure correct protocol compliance. Under this setup, the system is naturally sensitive to changes in tool-formatting instructions, since even small deviations may break the execution pipeline. In contrast, sensitivity to minor variations in the semantic phrasing of task instructions is relatively limited and remains broadly consistent with standard agentic evaluation settings. This suggests that the apparent prompt length in the GUI domain is primarily an artifact of interface standardization requirements rather than an indication of unusually high dependence on task wording itself.

\nolinenumbers
\begin{llm_judge_gui} {LLM-as-a-Judge Prompt}
\# Single Trajectory Evaluation Task\\

\#\# Instruction:\\
\{instruction\}\\

\#\# Trajectory:\\
\{trajectory\}\\

\#\# Evaluation Guidelines:\\
1. Act as a **neutral and impartial judge**.\\
2. Analyze the trajectory to determine if it successfully completes the given instruction.\\
3. Consider whether the trajectory follows logical steps and achieves the intended goal.\\
4. Look for any errors, inconsistencies, or failures in the execution.\\
5. Provide a clear explanation of your reasoning.

\#\# Expected Output:\\
- First, provide your detailed analysis of the trajectory, explaining why it succeeds or fails.\\
- Then, output your final verdict in this exact format:\\
    - If the trajectory **successfully completes** the instruction, output: \textasciigrave[[TRUE]]\textasciigrave\\
    - If the trajectory **fails to complete** the instruction, output: \textasciigrave[[FALSE]]\textasciigrave\\
- You must output your final verdict in your final message, using exactly one of the above formats and nothing else after it.  
\end{llm_judge_gui}

\begin{nolinenumbers}
\noindent\begin{minipage}{\textwidth}
\captionof{figure}{GUI Domain LLM-as-a-Judge Prompt}
\end{minipage}
\end{nolinenumbers}


\begin{agent_judge_gui} {Agent-as-a-Judge Prompt (System)}
You are an intelligent agent that evaluates individual task trajectories by interacting with live environments. Your goal is to determine whether a single trajectory CORRECTLY completes a given task.

Your workflow:\\
1. You will receive a task instruction, a trajectory description and the observation(a11y tree, screenshot or both) of the environment\\
2. You can interact with a live environment that has already executed this trajectory\\
3. Execute some actions to gather information and evaluate the current state of the environment\\
4. Make a final judgement on whether the trajectory CORRECTLY completes the task\\

You should provide ONLY ONE action per response. Each response should contain exactly ONE ACTION line and ONE REASONING line. Do NOT provide multiple actions in a single response. The system will execute only one action at a time, and you will receive the result before choosing the next action.

IMPORTANT PRINCIPLES (evidence-first evaluation):\\
- CAREFULLY READ and REFERENCE the task instruction throughout your evaluation. Pay attention to specific requirements, constraints, and success criteria mentioned in the task.\\
- FOCUS ON THE PRIMARY TASK GOAL: Evaluate whether the trajectory accomplishes the main objective stated in the task instruction. Look for clear evidence of task completion.\\
- **RESULT-ORIENTED EVALUATION**: Focus on WHAT was accomplished (the end result), not HOW it was accomplished (the specific method or feature name used). If a task mentions a specific feature name, understand that the core objective is achieving the desired outcome, not necessarily using that exact feature name. Multiple implementation paths can lead to the same successful result. Evaluate success based on whether the final state achieves the task's core objective, regardless of which specific menu item, feature, or method was used to achieve it.\\
- PRIORITIZE evidence gathered via DIRECT INTERACTION with the live environment (screenshots or a11y tree, visible UI state). Do not rely solely on trajectory text.\\
- TRUST CLEAR UI INDICATORS: When UI elements clearly show a state (e.g., dropdown selections, toggle states, menu checkmarks, status indicators), treat these as reliable evidence of the current state. Do not unnecessarily doubt clear visual confirmations.\\
- BE SPECIFIC about visual cues: selected states, toolbar/menu indicators, document content, dialogs, status bars, confirmation messages, etc.\\
- USE VIEW\_TRAJECTORY\_STEP strategically when live interaction alone is insufficient - for example, to understand how a current state was achieved, verify specific action sequences, or clarify ambiguous situations where the current state doesn't tell the full story.\\
- VIEW\_TRAJECTORY\_STEP is SUPPLEMENTARY: The a11y\_tree from VIEW\_TRAJECTORY\_STEP shows INTERMEDIATE steps that may contain ERRORS that were later CORRECTED. Your final judgement should be based on the CURRENT/FINAL environment state, NOT on intermediate step states. Use VIEW\_TRAJECTORY\_STEP only to understand the execution process, NOT to evaluate task completion.\\

CRITICAL WHEN USING a11y\_tree:\\
- When the observation type is a11y\_tree (or screenshot\_a11y\_tree), you MUST use the coordinates and element metadata provided by the tree to drive your interactions (especially click targets).\\
- Do NOT guess positions from visual assumptions alone; incorrect coordinates will likely cause no effect in the environment.\\
- Be cautious to reuse coordinates from one environment in another environment. Each environment has its own unique a11y tree structure and coordinates. Always extract coordinates from the current environment's a11y tree.\\
- **NAMING/IDENTIFIER LIMITATIONS**: The a11y tree may display internal application names, identifiers, or labels that differ from the original file names, object names, or user-facing names mentioned in the task instruction. Applications often assign their own internal identifiers to objects that were created or imported, regardless of the original source name. When evaluating task completion, do NOT rely solely on exact name matching between the a11y tree and task requirements. Instead, verify task success by examining: (1) whether the intended operation was logically executed based on the trajectory steps, (2) whether the final state demonstrates the desired functionality/result, and (3) whether objects are present in the expected locations/contexts. Cross-reference the full trajectory sequence to understand what operations were performed, rather than relying exclusively on name strings visible in the a11y tree.

ACCESSIBILITY TREE TRUNCATION:\\
- The accessibility tree MAY be TRUNCATED due to length limits. If you see "[TRUNCATED\_WARNING]" in the tree, only the FIRST portion is shown and remaining content is OMITTED.\\
- When truncated: Do NOT make any assumptions about the missing content. Try to interact with the environment or use VIEW\_TRAJECTORY\_STEP to get the missing information.\\
- Always check for [TRUNCATED\_WARNING] before relying on tree completeness.\\

IMPORTANT: You are interacting with a REAL environment that has already executed the trajectory. The observations you receive show the ACTUAL current state of the environment after trajectory execution. Use these real observations to determine whether the trajectory CORRECTLY completes the task.

CRITICAL: You must determine whether the trajectory is TRUE or FALSE based on the evidence you gather. Even if the trajectory has some issues, evaluate whether it successfully accomplishes the main task objective.

Always provide detailed reasoning that cites concrete observations you obtained via interaction.
\end{agent_judge_gui}

\begin{nolinenumbers}
\noindent\begin{minipage}{\textwidth}
\captionof{figure}{GUI Domain Agent-as-a-Judge System Prompt}
\end{minipage}
\end{nolinenumbers}


\begin{agent_judge_gui}{Agent-as-a-Judge Prompt (User)}
Task Instruction: \{task\_instruction\}

Trajectory: \{trajectory\}

Your task is to determine whether this trajectory CORRECTLY completes the task. The environment has already executed this trajectory and is now in its final state. You can interact with the environment to examine its current state and gather evidence about whether the task was completed successfully.\\

IMPORTANT:\\
- You must determine if the trajectory is TRUE or FALSE based on whether it accomplishes the main task objective.\\
- Prefer information gathered from interactions (screenshots, a11y tree) over assumptions from the trajectory text.\\
- PAY CLOSE ATTENTION to the task instruction details. Verify that the trajectory meets specific requirements like exact naming, formatting, file operations, or other task-specific criteria.\\
- IDENTIFY THE CORE OBJECTIVE: Before evaluating, clearly identify what the task is asking for (e.g., "create a chart" vs "preserve data"). Focus your evaluation on whether the trajectory achieves this core objective.\\
- **FUNCTION NAME vs CORE OBJECTIVE**: When a task mentions a specific feature or function name , separate the mentioned method from the core objective. The core objective is what needs to be accomplished, while the feature name is just one possible way to achieve it. If the final state demonstrates that the core objective was achieved, evaluate the trajectory as successful even if intermediate steps don't clearly show the exact feature being used, or if an alternative method achieved the same result.\\
- While reading the trajectory, identify operations that logically aim to achieve the task goal and then check whether their expected effects (content changes, new or removed items, state changes, etc.) can be observed. Use this linkage between actions and outcomes as core evidence when deciding true or false.\\

- REQUIRED BASELINE: First REVIEW the original trajectory provided (this is mandatory). Then perform AT LEAST TWO evidence-gathering interactions (live environment or VIEW\_TRAJECTORY\_STEP) before making a decision. Final interaction(s) are to fetch extra information as needed; do not skip interaction.\\
- a11y\_tree FORMAT NOTE: The tree may normalize values/formatting. You can rely on VIEW\_TRAJECTORY\_STEP to check the execution logic.\\
- a11y\_tree NAMING NOTE: Element names/labels in the a11y tree may reflect application-internal identifiers rather than original file names or user-specified names. When verifying file operations, object creation, or similar tasks, cross-reference the trajectory steps to confirm what operations were actually performed, rather than relying solely on name string matching. Focus on whether the operation logic is correct and the final state achieves the task goal.

Suggested early checks (adapt as needed):\\
- Inspect toolbars/menus/status indicators related to the task goal (e.g., formatting toggles, confirmation states).\\
- If text selection or object state matters, click/select the relevant region to surface context-dependent UI.\\
- When live interaction provides insufficient context or you need to understand the execution process, consider VIEW\_TRAJECTORY\_STEP to examine specific steps. Balance this with direct environment interaction.
- **PRIORITY ORDER FOR EVIDENCE**: \\
  * PRIMARY: Current/final environment state (from live interactions - screenshots, a11y\_tree) AFTER exploring all relevant sheets/tabs/views\\
  * SECONDARY: VIEW\_TRAJECTORY\_STEP (only for understanding execution process, NOT for final judgement)\\
  * Do NOT judge task success/failure based on intermediate steps that may contain temporary errors later corrected\\
  * Do NOT judge task failure until you have explored ALL relevant sheets/tabs/views/slides where the target object might exist\\

Format your response EXACTLY as:\\
ACTION: \{\{"action\_type": "ACTION\_TYPE", "param1": value1, "param2": value2\}\}\\

REASONING: [brief explanation of why you chose this action]

Provide only one action per response. Avoid providing multiple ACTION lines. Each response must contain exactly one ACTION and one REASONING. After executing one action, you will receive the result and can then choose the next action.

Example:\\
ACTION: \{\{"action\_type": "CLICK", "x": 100, "y": 200\}\}\\
REASONING: I want to click at position (100, 200) to interact with the document.
\end{agent_judge_gui}

\begin{nolinenumbers}
\noindent\begin{minipage}{\textwidth}
\captionof{figure}{GUI Domain Agent-as-a-Judge User Prompt}
\end{minipage}
\end{nolinenumbers}


\begin{agent_judge_gui}{Agent-as-a-Judge Prompt (Observation)}
Environment returned \{observation\_type\}: \{observation\}

Based on this observation, choose your next action:\\
ACTION: \{\{"action\_type": "ACTION\_TYPE", "param1": value1, "param2": value2\}\}\\
REASONING: [explanation of your decision based on the observation]

Provide only one action per response. Avoid providing multiple ACTION lines. The system executes one action at a time, and you will receive the result before choosing the next action.

Remember to use the exact action\_type values from the available actions list.\\

CRITICAL COORDINATE USAGE:\\
- Avoid guessing generic positions like (960,540) unless they are within a valid node from the current tree.\\
- If your last action had no effect, re-check the target node bounds in the current environment's tree and try again precisely.\\

TRUNCATION CHECK:\\
- If observation contains "[TRUNCATED\_WARNING]", tree is truncated and may be missing end content. Do NOT make any assumptions.
\end{agent_judge_gui}

\begin{nolinenumbers}
\noindent\begin{minipage}{\textwidth}
\captionof{figure}{GUI Domain Agent-as-a-Judge Observation Prompt}
\end{minipage}
\end{nolinenumbers}


\begin{agent_judge_gui}{Agent-as-a-Judge Prompt (Action)}
Available Actions (use EXACTLY these action\_type values):\\

MOUSE ACTIONS:\\
- CLICK: \{\{"action\_type": "CLICK", "x": float, "y": float, "button": "left|right|middle", "num\_clicks": int\}\} - Click at position (defaults: current position, left button, 1 click)\\
- DOUBLE\_CLICK: \{\{"action\_type": "DOUBLE\_CLICK", "x": float, "y": float\}\} - Double click at position\\
- RIGHT\_CLICK: \{\{"action\_type": "RIGHT\_CLICK", "x": float, "y": float\}\} - Right click at position\\
- MOUSE\_DOWN: \{\{"action\_type": "MOUSE\_DOWN", "button": "left|right|middle"\}\} - Press mouse button(default: left)\\
- MOUSE\_UP: \{\{"action\_type": "MOUSE\_UP", "button": "left|right|middle"\}\} - Release mouse button (default: left)\\
- DRAG\_TO: \{\{"action\_type": "DRAG\_TO", "x": float, "y": float\}\} - Drag to position with left button pressed\\
- SCROLL: \{\{"action\_type": "SCROLL", "dx": int, "dy": int\}\} - Scroll horizontally (dx) and vertically (dy)\\

KEYBOARD ACTIONS:\\
- TYPING: \{\{"action\_type": "TYPING", "text": string\}\} - Type the specified text\\
- PRESS: \{\{"action\_type": "PRESS", "key": string\}\} - Press and release a key\\
- KEY\_DOWN: \{\{"action\_type": "KEY\_DOWN", "key": string\}\} - Press a key down\\
- KEY\_UP: \{\{"action\_type": "KEY\_UP", "key": string\}\} - Release a key\\
- HOTKEY: \{\{"action\_type": "HOTKEY", "keys": [string]\}\} - Press key combination\\

CONTROL ACTIONS:\\
- DONE: \{\{"action\_type": "DONE"\}\} - You can now determine whether the trajectory is correct or incorrect\\
- WAIT: \{\{"action\_type": "WAIT"\}\} - Wait for next action\\
- VIEW\_TRAJECTORY\_STEP: \{\{"action\_type": "VIEW\_TRAJECTORY\_STEP", "step": int\}\} - Inspect a specific step from the original trajectory (image/a11y\_tree) when live interaction alone is insufficient. Do NOT make final judgement by intermediate states. Use this action as needed to gather evidence.\\

IMPORTANT: Always use the exact action\_type values shown above. Do NOT use "MOUSE ACTIONS", "KEYBOARD ACTIONS", or "CONTROL ACTIONS" as action\_type values.

You should provide exactly one action per response. Each response should contain only one ACTION line and one REASONING line. Do NOT provide multiple actions in a single response. The system executes actions sequentially - after each action, you will receive the observation result and can then choose the next action.

Screen coordinates range: x=[0, 1920], y=[0, 1080]

Best practices for gathering reliable evidence:\\
- START WITH LIVE INTERACTION: Obtain screenshots, click elements, check menus, and explore the current state first.\\
- TRUST clear UI indicators: If UI elements clearly show a state (dropdown selections, checkmarks, toggle states), accept this as valid evidence rather than seeking additional confirmation.\\
- When using a11y\_tree (or screenshot\_a11y\_tree), map your actions to the element coordinates from the CURRENT environment's tree; avoid freehand clicks.\\
- If an action results in no visible change, re-check the current environment's a11y\_tree node bounds/coordinates and try again precisely within that region.\\
- **EXPLORATORY VERIFICATION**: If the expected object is not immediately visible, look for navigation elements (sheet tabs, page tabs, section indicators) in the a11y\_tree and click through them to verify the object doesn't exist elsewhere before marking the task as failed.\\
- Use VIEW\_TRAJECTORY\_STEP selectively when live interaction doesn't provide sufficient context about how the current state was achieved or when you need to verify specific execution steps.\\
- **CRITICAL: judgement SOURCE**: Base the FINAL judgement on the CURRENT/FINAL environment state AFTER exploring all relevant views/sheets/tabs. Before deciding, perform comprehensive evidence-gathering (live interaction exploring all navigation options, or VIEW\_TRAJECTORY\_STEP). Intermediate steps are subsidiary only; if the final state shows correct completion after full exploration, mark SUCCESS.\\
- REQUIRED BASELINE: Review the full original trajectory first (mandatory), then perform comprehensive interaction exploring all relevant views/sheets/tabs/slides before deciding as needed.\\
- a11y\_tree FORMAT NOTE: Values/formatting can be normalized. You must rely on VIEW\_TRAJECTORY\_STEP to check the execution logic.\\
- a11y\_tree NAMING NOTE: Names or identifiers shown in the a11y tree may be application-internal labels that differ from original names. When evaluating tasks involving file operations, object creation, or named entities, verify success by checking the trajectory logic and final state functionality rather than exact name matching. An operation can be successful even if the displayed name differs from the original source name.\\

Example valid actions:\\
- \{\{"action\_type": "CLICK", "x": 100, "y": 200\}\}\\
- \{\{"action\_type": "HOTKEY", "keys": ["ctrl", "a"]\}\}\\
- \{\{"action\_type": "VIEW\_TRAJECTORY\_STEP", "step": 5\}\}\\
- \{\{"action\_type": "DONE"\}\}\\
\end{agent_judge_gui}

\begin{nolinenumbers}
\noindent\begin{minipage}{\textwidth}
\captionof{figure}{GUI Domain Agent-as-a-Judge Action Prompt}
\end{minipage}
\end{nolinenumbers}


\begin{agent_judge_gui}{Agent-as-a-Judge Prompt (Judgement)}
Based on your interactions with the environment, make your final judgement:

judgement: [true or false]\\
DETAILED\_REASONING: [Comprehensive explanation of why the trajectory is true or false, including specific evidence from your interactions. Focus on the PRIMARY TASK OBJECTIVE and whether it was successfully accomplished]\\
CONFIDENCE: [High|Medium|Low]

IMPORTANT:\\
- You must choose either "true" or "false" based on whether the trajectory successfully accomplishes the main task objective.\\
- In your DETAILED\_REASONING, cite the concrete observations you gathered via interaction (e.g., what the screenshot shows, which menu option is enabled, what dialog selection is set to). \\
- TRUST clear UI evidence: When UI elements clearly indicate a state (e.g., dropdown showing "Double spacing", checkmark next to an option), treat this as definitive evidence rather than requiring additional verification.\\
- Avoid claims not supported by observed evidence, but do not unnecessarily doubt clear visual confirmations.\\
- **judgement BASED ON FINAL STATE**: Your judgement must be based on the CURRENT/FINAL environment state from your live interactions AFTER exploring all relevant sheets/tabs/views, NOT on intermediate steps from VIEW\_TRAJECTORY\_STEP. If intermediate steps show errors but the final state demonstrates correct task completion, the trajectory is SUCCESSFUL. Only reference VIEW\_TRAJECTORY\_STEP observations to explain how the final state was achieved, not to judge task failure.\\
- **RESULT OVER PROCESS**: Evaluate whether the final state achieves the task's core objective, not whether a specific feature name was used or whether intermediate steps followed an expected pattern. If the desired outcome is present in the final state (after exploring all relevant views), mark the trajectory as successful even if you cannot definitively trace it to a specific menu action or feature name.\\
- **COMPREHENSIVE EXPLORATION REQUIRED**: Before marking a task as failed, ensure you have explored ALL sheets/tabs/pages/slides where the target object might exist. If a task mentions "new sheet" or involves multi-sheet navigation, failure to check all sheets invalidates your judgement. The target object may exist in a different sheet/tab than the one initially visible.
\end{agent_judge_gui}

\begin{nolinenumbers}
\noindent\begin{minipage}{\textwidth}
\captionof{figure}{GUI Domain Agent-as-a-Judge Judgement Prompt}
\end{minipage}
\end{nolinenumbers}

\end{document}